\documentclass{article} % For LaTeX2e
\usepackage{iclr2024_conference,times}

\usepackage{amsmath,amsfonts,bm}

\def\eqref#1{equation~\ref{#1}}
\def\1{\bm{1}}

\DeclareMathAlphabet{\mathsfit}{\encodingdefault}{\sfdefault}{m}{sl}
\SetMathAlphabet{\mathsfit}{bold}{\encodingdefault}{\sfdefault}{bx}{n}

\usepackage{hyperref}
\usepackage{url}

\usepackage{longtable}
\usepackage{booktabs}
\usepackage{forest}
\usepackage{xcolor}

\usepackage{geometry}
\usepackage{tikz}
\usetikzlibrary{trees,shapes.geometric,calc}

\title{A Survey of Prompt Engineering Methods in Large Language Models for Different NLP Tasks}

\author{Shubham Vatsal \& Harsh Dubey  \\
Department of Computer Science\\
New York University, CIMS\\
New York, USA \\
\texttt{\{sv2128,hd2225\}@nyu.edu}
}

\iclrfinalcopy % Uncomment for camera-ready version, but NOT for submission.
\begin{document}

\maketitle

\begin{abstract}
% Large language models (LLMs) have shown remarkable performance on many tasks in different domains including a few of the medical benchmarks.
Large language models (LLMs) have shown remarkable performance on many different Natural Language Processing (NLP) tasks.  Prompt engineering plays a key role in adding more to the already existing abilities of LLMs to achieve significant performance gains on various NLP tasks. Prompt engineering requires composing natural language instructions called prompts to elicit knowledge from LLMs in a structured way. Unlike previous state-of-the-art (SoTA) models, prompt engineering does not require extensive parameter re-training or fine-tuning based on the given NLP task and thus solely operates on the embedded knowledge of LLMs. Additionally, LLM enthusiasts can intelligently extract LLMs' knowledge through a basic natural language conversational exchange or prompt engineering, allowing more and more people even without deep mathematical machine learning background to experiment with LLMs. With prompt engineering gaining popularity in the last two years, researchers have come up with numerous engineering techniques around designing prompts to improve accuracy of information extraction from the LLMs. In this paper, we summarize different prompting techniques and club them together based on different NLP tasks that they have been used for. We further granularly highlight the performance of these prompting strategies on various datasets belonging to that NLP task, talk about the corresponding LLMs used, present a taxonomy diagram and discuss the possible SoTA for specific datasets. In total, we read and present a survey of 44 research papers which talk about 39 different prompting methods on 29 different NLP tasks of which most of them have been published in the last two years.
\end{abstract}

% \begin{table*}[h]
%     \centering
%     \begin{tabular}{lcccc}
%         \textbf{Dataset}     & \textbf{ProcessBank} & \textbf{BioMRC}  & \textbf{CliCR} & \textbf{MASH-QA}  \\
%         \midrule
%         \# QA Pairs & 150     & 6250     & 7184 & 3587\\
%         Avg Context Length    & 85     & 255   & 696  & 1461 \\
%         Max Context Length    & 266    & 510  & 3952 & 3952\\
%         % MASH-QA     & 35k    & XX     &   696   \\
%     \end{tabular}
%     \caption{Corpus Level Statistics}
%     \label{tab:datset}
% \end{table*}

\section{Introduction}

% About LLMs
Artificial Intelligence has advanced significantly with the introduction of LLMs. LLMs are trained on huge corpora of text documents with millions and billions of tokens. It has been shown that as the number of model parameters increase, the performance of machine learning models improve and such has been the case with these LLMs. They have attained unprecedented performance on a wide array of NLP tasks \cite{chang2023survey} because of which they have attracted a lot of interest from academia and different industries including medicine, law, finance and more. The present phase of research on LLMs focuses on their reasoning capacity via prompts rather than just next token prediction which has opened a new field of research around prompt engineering.

% About Prompt Engineering
Prompt engineering is the process of creating natural language instructions, or prompts, to extract knowledge from LLMs in an organized manner. Prompt engineering, in contrast to earlier conventional models, relies only on the embedded knowledge of LLMs and does not require extensive parameter re-training or fine-tuning based on the underlying NLP task. Understanding model parameters in terms of real world knowledge embedded in them is beyond human capabilities and hence this new field of prompt engineering has caught everyone's attention as it allows natural language exchange between researchers and LLMs to achieve the goals of the underlying NLP task.

% About our work
In this work, we enumerate several prompting strategies and group them according to different NLP tasks that they have been used for. We provide a taxonomy diagram, tabulate the prompting techniques tried on various datasets for different NLP tasks, discuss the LLMs employed, and list potential SoTA methods for each dataset. As a part of this survey, we have reviewed and analyzed 44 research papers in total, the majority of which have been published in the previous two years and cover 39 prompting techniques applied on 29 different NLP tasks. There have not been a lot of prior systematic surveys on prompt engineering. \citet{sahoo2024systematic} surveys 29 prompting technique papers based on their applications. This is a very broad categorization as a single application can encapsulate numerous NLP tasks. For example, one of the applications which they discuss is \textit{reasoning and logic} which can have plethora of NLP tasks like \textit{commonsense reasoning}, \textit{mathemathical problem solving}, \textit{multi-hop reasoning} etc. This is different from our approach as we take a more granular categorization of prompting strategies based on the NLP tasks. \citet{edemacu2024privacy} provides an overview of privacy protection prompting methods and thus focuses on a comparatively small sub-field of prompt engineering. \citet{chen2023unleashing} limits the discussion of prompting strategies to some 9-10 methodologies and also does not incorporate categorizing them based on the NLP tasks.

% Section descriptions

The rest of the paper is organized in the following way. Section 2 talks about various prompt engineering techniques and section 3 highlights different NLP tasks. The sub-sections of section 3 discuss different prompting strategies that have been applied on a given NLP task and their corresponding results. Section 4 concludes the paper.

\section{Prompt Engineering Techniques}
In this section, we talk briefly about different prompting methods and how they bring improvement in existing performance as and when they were published. An important thing to note here is that most of the following prompting strategies have been experimented in two different variations or settings if not more. These variations include zero-shot and few-shot. Some of the prompting techniques may inherently exist in either zero-shot or few-shot variation and there may not be a possibility for any other variation to exist. In zero-shot \cite{radford2019language} setting, there is no training data involved and an LLM is asked to perform a task through prompt instructions while completely relying on it's embedded knowledge learnt during it's pre-training phase. On the other hand, in few-shot variation \cite{brown2020language}, few training datapoints are provided along with task-based prompt instructions for better comprehension of the task. The results from various prompt engineering works have shown few-shot variations to have helped improve the performance but this comes at a cost of carefully preparing few-shot datapoints as the LLM can show unexplained bias towards the curated few-shot datapoints.
\subsection{Basic/Standard/Vanilla Prompting}
Basic prompting refers to the method of directly throwing a query at the LLM without any engineering around it to improve the LLM's performance which is the core goal behind most of the prompting strategies. Basic prompting also goes by the name of Standard or Vanilla prompting in different research papers.
\subsection{Chain-of-Thought (CoT)}
In this prompting strategy \cite{wei2022chain}, the authors build up on the idea of how human beings break a complex problem into smaller easier sub-problems before arriving at the final solution of the complex problem. Along similar lines, the authors investigate how capabilities of LLMs to do complicated reasoning is inherently enhanced by producing a chain of thought, or a sequence of intermediate reasoning steps. The results show a considerable improvement from Basic prompting with the maximum difference between CoT and Basic prompting results being as big as around 39\% for Mathematical Problem Solving task and around 26\% for Commonsense Reasoning task. This work opened a new direction of research for the field of prompt engineering.

\subsection{Self-Consistency}
Self-Consistency \cite{wang2022self} prompting technique is based on the intuition that complex reasoning problems can be solved in multiple ways and hence the correct answer can be reached via different reasoning paths. Self-Consistency uses a novel decoding strategy unlike the greedy one being used by CoT and consists of three important steps. The first step requires prompting the LLM using CoT, the second step samples diverse reasoning paths from LLM's decoder and the final step involves choosing the most consistent answer across multiple reasoning paths. Self-Consistency on an average achieves 11\% gain on Mathematical Problem Solving task, 3\% gain on Commonsense Reasoning task and 6\% gain on Multi-Hop Reasoning task when compared to CoT.

\subsection{Ensemble Refinement (ER)}
This prompting method has been discussed in \citet{singhal2023towards}. It builds on top of CoT and Self-Consistency. ER consists of two stages. First, given a few-shot CoT prompt and a query, LLM is made to produce multiple generations by adjusting it's temperature. Each generation contains a reasoning and an answer for the query. Next, the LLM is conditioned on the original prompt, query and the concatenated generations from the previous stage to generate a better explanation and an answer. This second stage is done multiple times followed by a majority voting over these second stage generated answers just as it is done in case of Self-Consistency to select the final answer. ER is seen to perform better than CoT and Self-Consistency across many datasets belonging to the Context-Free Question-Answering task.

\subsection{Automatic Chain-of-Thought (Auto-CoT)}
In this work \cite{zhang2022automatic}, the authors address the problem faced by few-shot CoT or manual CoT which is the need of curation of good quality training datapoints. Auto-CoT consists of two primary steps. The first one requires dividing queries of a given dataset into a few clusters. The second one involves choosing a representative query from each cluster and then generating its corresponding reasoning chain using zero-shot CoT. The authors claim that Auto-CoT either outperforms or matches the performance of few-shot CoT across Mathematical Problem Solving, Multi-Hop Reasoning and Commonsense Reasoning task. This indicates that the step of curation of training datapoints for few-shot or manual CoT can be ruled out.

\subsection{Complex CoT}
\citet{fu2022complexity} introduces a new prompting strategy which aims at choosing complex datapoint prompts over simpler ones. The complexity of a datapoint is defined here by the number of reasoning steps involved with it. The authors hypothesize that the LLMs' reasoning performance can increase if complex datapoints are used as in-context training examples as they already subsume simpler datapoints. Another important aspect of Complex CoT apart from using complex datapoints as training examples is that during decoding, just like Self-Consistency, out of N sampled reasoning chains the majority answer over the top K most complex chains is chosen as the final answer. There is one other baseline prompting method which has been introduced in this paper called Random CoT. In Random CoT, the datapoints are randomly sampled without adhering to their complexity. Complex CoT achieves on an average a gain of 5.3\% accuracy and up to 18\% accuracy improvement across various datasets of Mathematical Problem Solving, Commonsense Reasoning, Table-Based Mathematical Problem Solving and Multi-Hop Reasoning tasks.

\subsection{Program-of-Thoughts (PoT)}
The authors of \citet{chen2022program} build up on CoT but in contrast to CoT which uses LLMs to perform both reasoning and computation, PoT generates Python programs and thus relegates computation part to a Python interpreter. This work argues that reduced LLM responsibilities make it more accurate especially for numerical reasoning. PoT gets an average performance gain over CoT of around 12\% across Mathematical Problem Solving, Table-Based Mathematical Problem Solving, Contextual Question-Answering and Conversational Contextual Question-Answering tasks.

\subsection{Least-to-Most}

Least-to-Most \cite{zhou2022least} prompting technique tries to address the problem of CoT where CoT fails to accurately solve problems harder than the exemplars shown in the prompts. It consists of two stages. First, the LLM is prompted to decompose a given problem into sub-problems. Next, the LLM is prompted to solve the sub-problems in a sequential manner. The answer to any sub-problem depends on the answer of the previous sub-problem. The authors show that Least-to-Most prompting is able to significantly outperform CoT and Basic prompting methods on Commonsense Reasoning, Language-Based Task Completion, Mathematical Problem Solving and Contextual Question-Answering tasks.

\subsection{Chain-of-Symbol (CoS)}

CoS \cite{hu2023chain} builds up on the idea of CoT. In conventional CoT, the intermediate chain of reasoning steps are represented in natural language. While this approach has shown remarkable results in many cases, it can include incorrect or redundant information as well. The authors of this work present their hypothesis that spatial descriptions are hard to express in natural language thus making it difficult for LLMs to understand. Instead, expressing these relationships using symbols in word sequences can be a better form of representation for LLMs.
CoS achieves an improvement of up to 60.8\% accuracy for Spatial Question-Answering task.

\subsection{Structured Chain-of-Thought (SCoT)}
The intuition behind SCoT \cite{li2023structured} is that structuring intermediate reasoning steps using program structures like sequencing, branching and looping helps in more accurate code generation than having intermediate reasoning steps in natural language as we see in conventional CoT. The authors claim that the former approach more closely mimics a human developer’s thought process than the latter one and the same has been confirmed by the the final results as SCoT outperforms CoT by up to 13.79\% for the Code Generation task.

\subsection{Plan-and-Solve (PS)}
\citet{wang2023plan} discusses and tries to address three shortcomings of CoT which are calculation errors, missing-step errors and semantic misunderstanding errors. PS contains two components where the first one requires devising a plan to divide the entire problem into smaller sub-problems and the second one needs to carry out these sub-problems according to the plan. A better version of PS called PS+ adds more detailed instructions which helps in improving the quality of reasoning steps.  PS prompting method improves the accuracy over CoT by at least 5\% for almost all the datasets in Mathematical Problem Solving task in zero-shot setting. Similarly, for the Commonsense Reasoning task, it consistently outperforms CoT by at least 5\% in zero-shot setting whereas for the Multi-Hop Reasoning task it gets around 2\% better accuracy score.

\subsection{MathPrompter}
\citet{imani2023mathprompter} tries to address two key problems of CoT for Mathematical Problem Solving task: (1) lack of validity of steps followed by CoT for solving a problem; (2) how confident is an LLM in it's predictions. MathPrompter prompting strategy consists of 4 steps in total. (I) Given a query, the first step requires to generate an algebraic expression for the query which replaces the numerical values by variables. (II) Next, LLM is prompted to solve the query analytically either by deriving the algebraic expression or writing a Python function. (III) Third, the query in step (I) is solved by assigning different values to the variables. (IV) If the solutions in (III) are correct over N iterations, the variables are finally replaced with original query values and the answer is computed. If not, then the steps (II), (III) and (IV) are repeated. MathPrompter is able to improve the performance on a dataset belonging to Mathematical Problem Solving task from 78.7\% to 92.5\%.

\subsection{Contrastive CoT/ Contrastive Self-Consistency}
The authors of \citet{chia2023contrastive} claim that Contrastive CoT or Contrastive Self Consistency is a general enhancement of CoT or Self-Consistency. The inspiration for this prompting approach is based on how humans can learn from both positive as well as negative examples. Along similar lines, in this prompting technique, both positive and negative demonstrations are provided to enhance the reasoning capabilities of the LLM. Contrastive CoT on an average is able to gain an average of 10\% improvement over conventional CoT for Mathematical Problem Solving task across multiple datasets. Similarly, Contrastive Self-Consistency is able to outperform conventional Self-Consistency by over 15\% for Mathematical Problem Solving task across multiple datasets. For Multi-Hop Reasoning task, both Contrastive CoT and Contrastive Self-Consistency have more than 10\% gains over their conventional counterparts.

\subsection{Federated Same/Different Parameter Self-Consistency/CoT (Fed-SP/DP-SC/CoT)} Introduced in \citet{liu2023federated}, this prompting method is based on the core idea of improving the reasoning capabilities of LLMs by using synonymous crowd-sourced queries. There are two slightly different variations of this prompting method. The first one is Fed-SP-SC where the crowd-sourced queries are paraphrased versions of the original query but with same parameters. Parameters here can refer to the numeric values in Mathematical Problem Solving task datapoints. For Fed-SP-SC, the answers are directly generated first and then Self-Consistency is applied on top of it. The other one is Fed-DP-CoT. In Fed-DP-CoT, LLMs are used to first generate answers to different queries and then they are federated by forming CoT to provide hints to the LLMs. The results for these methods on Mathematical Problem Solving task show that they are able to do better than conventional CoT by at least 10\% and up to 20\%. 

\subsection{Analogical Reasoning}
The authors of this work \cite{yasunaga2023large} draw their inspiration from a psychological notion, analogical reasoning, where people use pertinent prior experiences to solve new problems. In the realm of LLMs, the authors first prompt them to generate examples similar to that of the original problem followed by solving them and then proceed to answer the original problem. The results show that Analogical Reasoning is able to achieve an average accuracy gain of 4\% when compared to CoT across Mathematical Problem Solving, Code Generation, Logical Reasoning and Commonsense Reasoning tasks.

\subsection{Synthetic Prompting}
The authors of \citet{shao2023synthetic} come up with Synthetic prompting using LLMs to generate synthetic examples which are augmented to the existing hand-crafted examples as seen in a conventional few-shot setting. This prompting method involves two steps: (1) the backward step, where the LLM synthesizes a query based on a self-generated reasoning chain; and (2) the forward step, where the LLM generates a reasoning chain for the synthesized query, making the reasoning chain to be more accurate. Finally, to choose the best examples, this work uses an in-cluster complexity and the most complex examples with the longest reasoning chains are used during inference. The results show Synthetic prompting achieving up to 15.6\% absolute gains when experimented with different Mathematical Problem Solving, Commonsense Reasoning and Logical Reasoning task datasets.

\subsection{Tree-of-Thoughts (ToT)}

ToT \cite{yao2024tree} prompting technique has been drawn from the idea that any kind of problem solving requires searching through a combinatorial space represented as a tree where each node represents a partial solution and each branch corresponds to an operator that modifies it. Now, the decision about which branch to choose is determined by heuristics that help to navigate the problem-space and guide the problem-solver towards a solution. Based on this idea, the authors propose ToT which actively maintains a tree of thoughts where each thought is a coherent language sequence that serves as an intermediate reasoning step toward problem solving. This framework allows LLMs to evaluate the progress generated by thoughts while trying to solve the problem. ToT further incorporates search techniques such as breadth-first or depth-first search with the model's ability to generate and evaluate thoughts. ToT achieves 65\% better success rate than CoT on Mathematical Problem Solving task and around 40\% better success rate on different Logical Reasoning task datasets. ToT further achieves coherency score of 7.56 where CoT gets only 6.93 on an average on Free Response task.

\subsection{Logical Thoughts (LoT)}

In this work \cite{zhao2023enhancing}, the authors investigate the usage of logical equivalence in order to improve the zero-shot reasoning abilities of an LLM. In addition to allowing the LLM to reason step-by-step, LoT also allows the LLM to verify step-by-step in accordance with the guidelines provided by the \textit{Reductio ad Absurdum} principle and, if needed, amend the reasoning chain to ensure a valid inference. LoT is able to surpass CoT in Mathematical Problem Solving task by a maximum of 3.7\%, Commonsense Reasoning task by a maximum of 16.2\%, Logical Reasoning task by a maximum of 2.5\%, Causal Reasoning task by a maximum of 15.8\% and Social Reasoning task by a maximum of 10\% accuracy.

\subsection{Maieutic Prompting}
By using deep recursive reasoning to elicit abductive explanations for a variety of hypotheses, Maieutic prompting \cite{jung2022maieutic} encourages the LLM to produce consistent responses by collaboratively eliminating alternatives that contradict one another. The generation process of Maieutic prompting derives a tree structure of generated propositions, where one proposition establishes a logical ground for the correctness of one another. Finally, to infer the answer to the original query, the degree to which the LLM believes each proposition and the logical connections between propositions in the maieutic tree is measured. The results for Maieutic prompting on Commonsense Reasoning task shows that it is able to achieve up to 20\% better accuracy when compared to Basic prompting, CoT, Self-Consistency and GKP \cite{liu2021generated} while performing competitively with supervised models.

\subsection{Verify-and-Edit (VE)} \citet{zhao2023verify} focuses on developing a technique which can post-edit the reasoning chains generated by CoT for more factually aligned outputs. This method consists of three stages: (1) the deciding when to edit stage where the authors use Self-Consistency to find uncertain outputs; (2) the how to edit rationales stage where the authors edit CoT reasoning chains of uncertain outputs by searching for supporting facts from external knowledge sources and (3) the reasoning stage where the edited rationales from previous stage are used to come up with final answers. VE is able to outperform CoT, Self-Consistency and Basic prompting by up to 10\%  on Multi-Hop Reasoning task and by up to 2\% on Truthfulness task.

\subsection{Reason + Act (ReAct)}
\citet{yao2022react} presents ReAct, which combines reasoning and acting with LLMs to solve diverse language reasoning and decision making tasks. In order to enable the model to perform dynamic reasoning to build and modify high-level plans for acting (reason to act), ReAct prompts LLMs to generate verbal reasoning traces and actions related to a task in an interleaved manner. Another prompting method similar to ReAct discussed in \citet{yao2022react} is Act which basically removes thoughts or reasoning in ReAct trajectories but performs suboptimal to ReAct in all the discussed tasks. For Multi-Hop Reasoning and Truthfulness tasks, ReAct is able perform better than Basic prompting while being competitive with CoT. When ReAct is combined with CoT or Self-Consistency, it is able to get better results than CoT. For Language-Based Task Completion task, ReAct outperforms reinforcement learning methods with an absolute improvement of more than 10\% in success rates individually on different datasets.

\subsection{Active-Prompt}
\citet{diao2023active} proposes Active-Prompt to help LLMs adapt to different tasks with task-specific examples by identifying the most relevant datapoints to be used as examples while prompting the LLM in a few-shot setting. Active-Prompt is a four-step technique. In the first step, the LLM is prompted k times for each query in the training set to generate k possible answers with their corresponding reasoning chains. The next step requires calculating the uncertainty metric based on the answers generated in step one. In the third step, the top n most uncertain queries are selected and annotated by humans. In the final step, the new annotated examples are used to do few-shot prompting for the test data. The authors also introduce a different version of Active-Prompt called Random CoT where in step 3, top n queries are selected randomly than based on the uncertainty metric. The results show that Active-Prompt is able to get better results than Self-Consistency, CoT, Auto-CoT and Random CoT across multiple datasets for Mathematical Problem Solving, Commonsense Reasoning, Multi-Hop Reasoning, Commonsense Reasoning tasks.

\subsection{Thread-of-Thought (ThoT)}
\citet{zhou2023thread} proposes a prompting method focusing on handling long chaotic contexts. It is based on the idea that there is an unbroken flow of thought that people retain when going through a large amount of information, enabling the selective extraction of pertinent data and the rejection of irrelevant ones. This balance of attention across a document’s sections is important for accurate interpretation and response to the information supplied. ThoT consists of two steps. The first one requires the LLM to analyze and summarize the different sections of the context. In the second step, the LLM is prompted to answer the asked query based on the output of first step. ThoT is able to outperform CoT and Basic promoting techniques by achieving a score of around 0.56 exact match in Context-Free Question-Answering task. For Dialogue System task, ThoT is able to get the highest average score of 3.8 again surpassing other discussed prompting techniques.

\subsection{Implicit Retrieval Augmented Generation (Implicit RAG)}
Contrary to the conventional RAG \cite{lewis2020retrieval}, Implicit RAG  \cite{vatsal2024can,vatsal2024canpa} asks the LLM itself to retrieve important chunks or sections from the given context and then proceed to answer the asked query. This technique requires tuning of two hyper-parameters. The first one is the number of sections to extract whereas the second one is the number of words in each section. Implicit RAG achieves SoTA result on Contextual Question-Answering task in \citet{vatsal2024canpa} on Patient Case Reports dataset whereas achieved either SoTA or close to SoTA results on biomedical Contextual Question-Answering task datasets in \citet{vatsal2024can}.

\subsection{System 2 Attention (S2A)}
LLMs can often end up making erroneous judgments when presented with irrelevant context. \citet{weston2023system} tries to address this issue with two-step prompting strategy. The first step instructs the LLM to regenerate a given context such that the regenerated version does not contain any irrelevant parts that could adversely affect the output. The second step then instructs the LLM to produce the final response using the regenerated context from step 1. The results show that S2A is able to outperform Basic, CoT as well Instructed prompting \cite{shi2023large}  over different Truthfulness task datasets.

\subsection{Instructed Prompting}
Intructed prompting \cite{shi2023large} again revolves around the same idea as that of S2A which tries to address the issue of LLMs getting distracted by irrelevant context. It consists of only one step of explicitly instructing the language model to ignore irrelevant information in the problem description. Instructed prompting is able to achieve 88.2 normalized micro accuracy for Truthfulness task and is able to surpass all it's counterparts including CoT, Least-To-Most, Program prompting and Self-Consistency. Program prompting \cite{chowdhery2023palm} strategy here tries to solve a problem by writing a Python program for it. Later, the correctness of the written program is verified by running the Python code using an external Python interpreter to obtain the final answer.

\subsection{Chain-of-Verification (CoVe)}

LLMs are prone to generating factually incorrect information called hallucination. The authors of \citet{dhuliawala2023chain} try to address this problem of hallucination and improve performance via CoVe. CoVe performs four core steps. First, the LLM generates a baseline response for a given query. Second, using both the original query and the baseline response from step one, generate a list of verification queries that are capable of checking if there are any errors in the baseline response. Third, generate answers to all the verification queries from step three. Fourth, correct all the mistakes in the baseline response detected after step three and produce a revised response. The results show that CoVe is able to outperform CoT and Basic prompting by around at least 10\% on Context-Free Question-Answering, Contextual Question-Answering and Free Response tasks.

\subsection{Chain-of-Knowledge (CoK)}
Similar to CoVe, CoK \cite{li2023chaink} tries to address the issue of hallucination to get more accurate results. It's a three-stage prompting technique. The first stage is reasoning preparation where given a query, CoK prepares several preliminary rationales and answers while identifying the relevant knowledge domains. The second stage is dynamic knowledge adaptation where if there is no majority consensus among the answers, CoK corrects the rationales step by step by adapting knowledge from the identified domains in stage one. The third stage is answer consolidation which uses these corrected rationales from stage two to serve as a better foundation for the final answer consolidation. CoVe surpasses CoT, Self-Consistency,  VE and Basic prompting across Context-Free Question-Answering, Table-Based Question-Answering, Multi-Hop Reasoning and Truthfulness tasks and shows an improvement of at least 3\%, 3\%, 1\% and 1\% respectively.

\subsection{Chain-of-Code (CoC)}
In this work \cite{li2023chain}, the authors propose an extension to make LLM's code-oriented reasoning better. Here, the LLM not only writes a code for a program but also selectively simulates the interpreter by producing the expected outputs of certain lines of code which cannot be actually executed by an interpreter. The main idea is to motivate LLMs to format semantic sub-tasks in a program as flexible pseudocode that may be explicitly caught and passed off to an LLM for emulation at runtime which the authors call an LMulator. Experiments demonstrate CoC surpassing CoT and other baselines across a variety of tasks including Recommender System, Causal Reasoning, Commonsense Reasoning, Spatial Question-Answering, Emotion/Sentiment Understanding, Machine Translation, Logical Reasoning, Table-Based Mathematical Problem Solving and Mathematical Problem Solving.

\subsection{Program-aided Language Models (PAL)}
\citet{gao2023pal} proposes a prompting strategy that uses an LLM to read natural language problems and generate interleaved natural language and programming language statements as reasoning steps. Finally, a Python interpreter is used to execute programming statements to get the answer. The results show that PAL easily performs better than it's counterparts like CoT and Basic prompting across multiple NLP tasks including Mathematical Problem Solving, Table-Based Mathematical Problem Solving, Commonsense Reasoning and Logical Reasoning.

\subsection{Binder} 
\label{binder}

The authors claim Binder \cite{cheng2022binding} to be a training-free neural-symbolic technique that maps an input to a program which (I) enables binding of a single API of LLM functionalities to a programming language such as Python or SQL in order to increase it's coverage of grammar and to address a wider range of queries; (II) uses an LLM as the underlying model as well as the program parser during execution; (III) needs only a few in-context sample annotations. The binder pipeline has two stages. First, in the parsing stage, the LLM maps the input to a program given the query and knowledge sources. Second, in the execution stage, the LLM returns values in the chosen programming language and finally the program is run using an interpreter. Binder is able to get better accuracy when compared to previous methodologies which required explicit training or fine-tuning for Table-Based Truthfulness and Table-Based Question-Answering tasks.

\subsection{Dater} 

\citet{ye2023large} explores the idea of few-shot learning with LLMs to decompose evidence and queries for efficient table-based reasoning. This prompting strategy involves three important steps. It starts with decomposing a huge table into relevant smaller sub-tables given the query. Next, SQL programming language is used to decompose the complex natural language query into logical and numerical computations. Finally, the sub-tables and sub-queries from previous two steps are used to arrive at the final answer in a few-shot setting. The results show that Dater is able to surpass previous methodologies which required explicit fine-tuning by at least 2\% in Table-Based Truthfulness task. Similarly, for Table-Based Question-Answering task,  it is able to outperform such methods by at least 1\%. Dater is also able to do better than Binder for both the above-mentioned tasks.

\subsection{Chain-of-Table}
In \citet{wang2024chain}, the authors build up on the famous prompting technique of CoT and bring it to the tabular setting. This multi-step tabular prompting approach leads to more accurate table understanding. Chain-of-Table is a three-step prompting technique. The first step instructs the LLM to dynamically plan the next table operation by in-context learning. An operation here could be anything from addition of columns to sorting of rows. The second step generates arguments for the selected table operation. The first two steps help in transforming the table and creating various intermediate table representations with the goal of answering the original query. In the final step, the last table representation from the first two steps is used to finally answer the query. Chain-of-Table achieves SoTA performance on Table-Based Question-Answering and Table-Based Truthfulness tasks. For Table-Based Question-Answering task, it gets around 3\% of average better performance whereas for Table-Based Truthfulness task it is able to get around 1.5\% of average better performance when compared to the prior SoTA results.

% \subsection{Prompt Policy Gradient (PromptPG)}

% \citet{lu2022dynamic} proposes PromptPG prompting strategy that learns to select in-context examples from a small amount of training data via policy gradient for prompt learning.

% Experimental results show that PromptPG method outperforms the best baseline by 5.31\% in terms of accuracy for Table-Based Mathematical Problem Solving task and reduces the prediction variance considerably compared to random selection.

\subsection{Decomposed Prompting (DecomP)}
\citet{khot2022decomposed} comes up with DecomP technique which decomposes a complex problem into simpler sub-problems and then delegates these to sub-problem specific LLMs, which have their own prompts and decomposers to further decompose the sub-problems. The decomposers can either resort to hierarchical decomposition, recursive decomposition or make external API calls to solve the sub-problem. DecomP is able to outperform CoT and Least-to-Most on an average by 25\% in terms of exact match for Commonsense Reasoning task. For Multi-Hop Reasoning task, DecomP is comfortably able to do better than CoT on four different datasets.
% \subsection{Directional Stimulus Prompting (DSP)}
% \cite{li2024guiding} 

\subsection{Three-Hop Reasoning (THOR)}

The authors of \citet{fei2023reasoning} come up with THOR to mimic human-like reasoning process for Emotion/Sentiment Understanding task. THOR consists of three steps. In the first step, the LLM is asked to identify the aspect mentioned in the given query. Next, based on previous step output and the original query, the LLM is asked to answer in detail about the underlying opinion embedded in the query. Finally, all of the above information is combined and the LLM is asked to infer the sentiment polarity associated with the given query. THOR is able to significantly surpass prior SoTA supervised as well as zero-shot models on multiple Emotion/Sentiment Understanding task datasets.

\subsection{Metacognitive Prompting (MP)}
MP \cite{wang2023metacognitive} is based on the concept of meta-cognition which is derived from cognitive psychology and relates to an individual’s awareness and self-reflection on their cognitive processes. It consists of five stages. 1) understanding the input text, 2) making a preliminary judgment, 3) critically evaluating this preliminary analysis, 4) reaching a final decision accompanied by an explanation of the reasoning, and 5) evaluating the confidence level in the entire process. The results show that MP consistently excels CoT and PS across numerous NLP tasks including Paraphrasing, Natural Language Inference, Contextual Question-Answering, Word Sense Disambiguation, Named Entity Recognition, Relation Extraction and Multilabel Text Classification.

\subsection{Chain-of-Event (CoE)}

\citet{bao2024chain} proposes CoE for the Summarization task. CoE has four sequential steps. The first one focuses on specific event extraction. Next, the events extracted in step one are analyzed and generalized into more concise and refined form. Third, the events generalized in the previous step are filtered and only those are selected which cover most of the text. In the last step, the events selected in step three are integrated based on their chronological order of importance. The results show that CoE is able to perform better than CoT across two Summarization datasets in terns of rouge score while also being more concise.

\subsection{Basic with Term Definitions}
This is one of prompting methods discussed in \citet{vatsal2024canpa}. In this method, basic prompt instructions get enhanced by addition of medical term definitions based on the hypothesis that adding these definitions would help the LLM in gaining more context while answering the asked query. But the results show that these term definitions do not really help possibly because of their narrow knowledge scope which may be conflicting with the bigger knowledge base of the LLM. Also, the definition of medical terms can change with the change in context and thus having a fixed definition for medical terms do not really help the LLMs and in return end up confusing them.

\subsection{Basic + Annotation Guideline-Based Prompting + Error Analysis-Based Prompting} \citet{hu2024improving} tests LLM capabilities in clinical Named Entity Recognition task. This prompting strategy has three different components. The basic component tells the LLM about the rudimentary information regarding the task and in what format the LLM should output the results. The annotation guideline component contains entity definitions and linguistic rules derived from the annotation guidelines. The error analysis component incorporates additional instructions following error analysis of LLM outputs using the training data. Different versions of this prompting method  have been also experimented by the authors by creating different combination of above-mentioned components. This prompting method is able to get on an average 0.57 exact match F1 score on multiple datasets belonging to Named Entity Recognition task.

\section{Prompt Engineering on Different NLP Tasks}
Different research papers have used different measures when it comes to categorizing a dataset under an NLP task and it keeps varying from one work to another. In this section, we try to standardize this and put a structure around these prior ways of categorization by defining different NLP tasks and putting different datasets under these tasks. We further talk about various prompting methods that have been used for these tasks. A taxonomy diagram reflecting this can be seen in Figure \ref{taxonomy}.  An important thing to note here is that it is quite possible that a dataset can belong to different NLP tasks at the same time. But that can result in a complex entanglement of structured analyses of how prompting techniques perform across various NLP tasks. Therefore, in our work, we make sure that a dataset belongs to only one NLP task to which it most strongly associates with. The following sub-sections each define a different NLP task, corresponding datasets and various prompting strategies that have been applied to those datasets. They further contain the potential SoTA prompting technique for each dataset. The performance of a prompting method varies based on the LLM used. Therefore, we have also included a list of LLMs that were used along with prompting strategies on a given dataset. For the SoTA, we have only mentioned the name of the prompting method, as in many cases a particular LLM has not been experimented with a given prompting method, making it unclear if it could have achieved SoTA performance. Hence, if any LLM from the list of LLMs, along with a prompting strategy, has been used to experiment with the given dataset and achieved the best performance, we have designated that as the SoTA regardless of the exact LLM used for that technique. Similarly, we haven't mentioned the evaluation metric while listing down the SoTA because they can differ across different

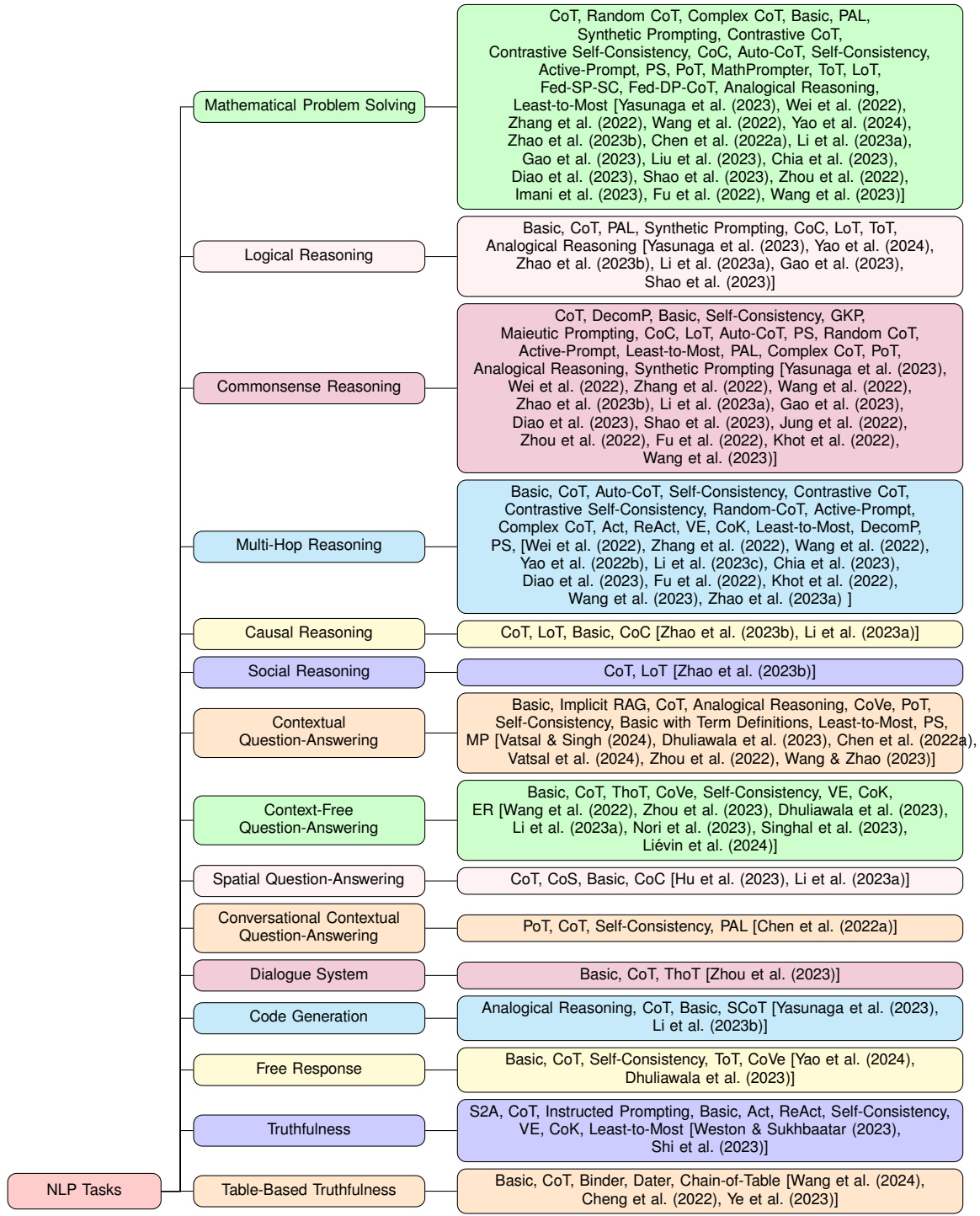
\begin{forest}
for tree={
    grow=east,
    draw,
    rounded corners,
    align=center,
    text width=4cm,
    inner xsep=4pt,
    inner ysep=2pt,
    l sep=5mm,
    s sep=1mm,
    parent anchor=east,
    child anchor=west,
    anchor=west,
    calign=first,
    edge path={
        \noexpand\path [draw, \forestoption{edge}] (!u.parent anchor) -- +(3mm,0) |- (.child anchor)\forestoption{edge label};
    },
    font=\sffamily\scriptsize,
    if level=0{
            fill=red!20,
            text width=2cm, % Different text width for level 0
            inner xsep=6pt, % Different inner xsep for level 0
            inner ysep=4pt,  % Different inner ysep for level 0
            align=center,
            text centered
        }{},
    if level=1{
            text width=3.3cm, % Different text width for level 1
            inner xsep=5pt, % Different inner xsep for level 1
            inner ysep=3pt,  % Different inner ysep for level 1
            align=center,
            text centered
        }{},
    if level=2{
            text width=7.7cm, % Different text width for level 2
            inner xsep=4pt, % Different inner xsep for level 2
            inner ysep=2pt,  % Different inner ysep for level 2
            align=center,
            text centered
        }{}
}
[NLP Tasks
    [Table-Based Truthfulness, fill=orange!20
        [{
        Basic, CoT, Binder, Dater, Chain-of-Table [\cite{wang2024chain}, \\ \cite{cheng2022binding}, \cite{ye2023large}]
        }, fill=orange!20]
    ]
    [Truthfulness, fill=blue!20
        [{
        S2A, CoT, Instructed Prompting, Basic, Act, ReAct, Self-Consistency, \\
        VE, CoK, Least-to-Most [\cite{weston2023system}, \\ \cite{shi2023large}]
        }, fill=blue!20]
    ]
    [Free Response, fill=yellow!20
        [{
        Basic, CoT, Self-Consistency, ToT, CoVe [\cite{yao2024tree}, \\
        \cite{dhuliawala2023chain}]
        }, fill=yellow!20]
    ]
    [Code Generation, fill=cyan!20
        [{
        Analogical Reasoning, CoT, Basic, SCoT [\cite{yasunaga2023large}, \\ \cite{li2023structured}]
        }, fill=cyan!20]
    ]
    [Dialogue System, fill=purple!20
        [{
        Basic, CoT, ThoT [\cite{zhou2023thread}]
        }, fill=purple!20]
    ]
    [Conversational Contextual \\ Question-Answering, fill=orange!20
        [{PoT, CoT, Self-Consistency, PAL [\cite{chen2022program}]}, fill=orange!20]
    ]
    [Spatial Question-Answering, fill=pink!20
        [{CoT, CoS, Basic, CoC [\cite{hu2023chain}, \cite{li2023chain}]}, fill=pink!20]
    ]
    [Context-Free \\ Question-Answering, fill=green!20
        [{Basic, CoT, ThoT, CoVe, Self-Consistency, VE, CoK, \\
        ER [\cite{wang2022self}, \cite{zhou2023thread}, \cite{dhuliawala2023chain}, \\ \cite{li2023chain}, \cite{nori2023capabilities}, \cite{singhal2023towards}, \\ \cite{lievin2024can}]}, fill=green!20]
    ]
    [Contextual \\ Question-Answering, fill=orange!20
        [{Basic, Implicit RAG, CoT, Analogical Reasoning, CoVe, PoT, \\
        Self-Consistency, Basic with Term Definitions,  Least-to-Most, PS, \\ MP [\cite{vatsal2024can}, \cite{dhuliawala2023chain}, \cite{chen2022program}, \\ \cite{vatsal2024canpa}, \cite{zhou2022least}, \cite{wang2023metacognitive}]}, fill=orange!20]
    ]
    [Social Reasoning, fill=blue!20
        [{CoT, LoT [\cite{zhao2023enhancing}]}, fill=blue!20]
    ]
    [Causal Reasoning, fill=yellow!20
        [{CoT, LoT, Basic, CoC [\cite{zhao2023enhancing}, \cite{li2023chain}]}, fill=yellow!20]
    ]
    [Multi-Hop Reasoning, fill=cyan!20
        [{Basic, CoT, Auto-CoT, Self-Consistency,
        Contrastive CoT, \\ Contrastive Self-Consistency,
        Random-CoT, Active-Prompt, \\ Complex CoT,
        Act, ReAct, VE, CoK, Least-to-Most,
        DecomP, \\
        PS, [\cite{wei2022chain},  
        \cite{zhang2022automatic}, \cite{wang2022self}, \\
        \cite{yao2022react}, \cite{li2023chaink},
        \cite{chia2023contrastive}, \\ \cite{diao2023active},  
        \cite{fu2022complexity},  \cite{khot2022decomposed}, \\ 
        \cite{wang2023plan}, \cite{zhao2023verify} ]}, fill=cyan!20]
    ]
    [Commonsense Reasoning, fill=purple!20
        [{CoT, DecomP, Basic, Self-Consistency, GKP, \\
        Maieutic Prompting, CoC, LoT, Auto-CoT, PS, Random CoT, \\ Active-Prompt, Least-to-Most, PAL, Complex CoT, PoT, \\ Analogical Reasoning, Synthetic Prompting [\cite{yasunaga2023large}, \\ \cite{wei2022chain}, \cite{zhang2022automatic}, \cite{wang2022self}, \\ \cite{zhao2023enhancing}, \cite{li2023chain},
        \cite{gao2023pal}, \\ \cite{diao2023active},
        \cite{shao2023synthetic},  \cite{jung2022maieutic}, \\  \cite{zhou2022least},
        \cite{fu2022complexity}, \cite{khot2022decomposed}, \\ \cite{wang2023plan}]}, fill=purple!20]
    ]
    [Logical Reasoning, fill=pink!20
        [{Basic, CoT, PAL, Synthetic Prompting, CoC, LoT, ToT, \\
        Analogical Reasoning [\cite{yasunaga2023large}, \cite{yao2024tree}, \\ \cite{zhao2023enhancing}, \cite{li2023chain}, \cite{gao2023pal}, \\
        \cite{shao2023synthetic}]}, fill=pink!20]
    ]
    [Mathematical Problem Solving, fill=green!20
        [{CoT, Random CoT, Complex CoT, Basic, PAL, \\
        Synthetic Prompting, Contrastive CoT, \\ Contrastive Self-Consistency, 
        CoC, Auto-CoT, Self-Consistency, \\ 
        Active-Prompt, PS, PoT, MathPrompter, ToT, LoT, \\
        Fed-SP-SC, Fed-DP-CoT, 
        Analogical Reasoning, \\
        Least-to-Most
        [\cite{yasunaga2023large},
        \cite{wei2022chain},\\
        \cite{zhang2022automatic}, \cite{wang2022self}, \cite{yao2024tree}, \\
        \cite{zhao2023enhancing},
        \cite{chen2022program}, \cite{li2023chain},\\
        \cite{gao2023pal},
        \cite{liu2023federated}, \cite{chia2023contrastive}, \\ \cite{diao2023active}, 
        \cite{shao2023synthetic},
        \cite{zhou2022least}, \\
        \cite{imani2023mathprompter},
        \cite{fu2022complexity}, \cite{wang2023plan}]}, fill=green!20]
    ]
]
\label{taxonomy}
\end{forest}

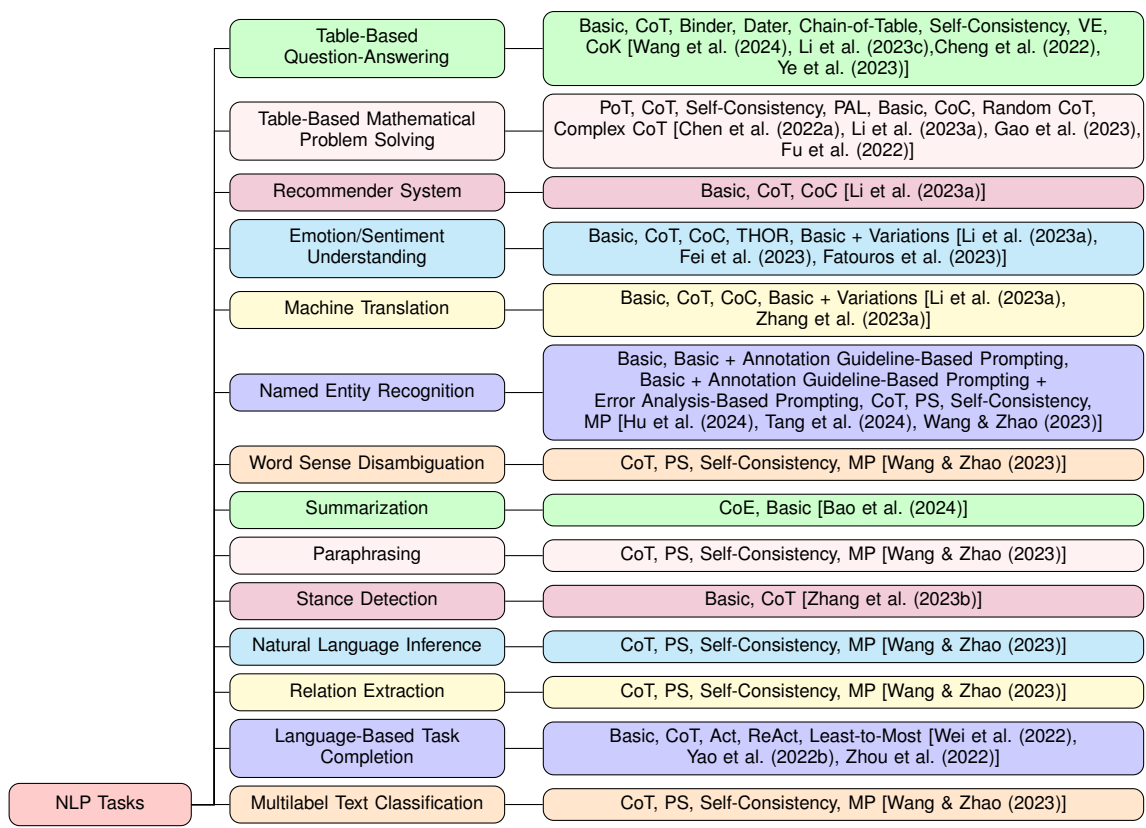
\begin{figure}[!ht]

\begin{forest}
for tree={
    grow=east,
    draw,
    rounded corners,
    align=center,
    text width=4cm,
    inner xsep=4pt,
    inner ysep=2pt,
    l sep=5mm,
    s sep=1mm,
    parent anchor=east,
    child anchor=west,
    anchor=west,
    calign=first,
    edge path={
        \noexpand\path [draw, \forestoption{edge}] (!u.parent anchor) -- +(3mm,0) |- (.child anchor)\forestoption{edge label};
    },
    font=\sffamily\scriptsize,
    if level=0{
            fill=red!20,
            text width=2cm, % Different text width for level 0
            inner xsep=6pt, % Different inner xsep for level 0
            inner ysep=4pt,  % Different inner ysep for level 0
            align=center,
            text centered
        }{},
    if level=1{
            text width=3.3cm, % Different text width for level 1
            inner xsep=5pt, % Different inner xsep for level 1
            inner ysep=3pt,  % Different inner ysep for level 1
            align=center,
            text centered
        }{},
    if level=2{
            text width=7.7cm, % Different text width for level 2
            inner xsep=4pt, % Different inner xsep for level 2
            inner ysep=2pt,  % Different inner ysep for level 2
            align=center,
            text centered
        }{}
}
[NLP Tasks
    [Multilabel Text Classification, fill=orange!20
        [{
        CoT, PS, Self-Consistency, MP [\cite{wang2023metacognitive}]
        }, fill=orange!20]
    ]
    [Language-Based Task \\ Completion, fill=blue!20
        [{
        Basic, CoT, Act, ReAct, Least-to-Most
         [\cite{wei2022chain},\\ \cite{yao2022react}, \cite{zhou2022least}]
        }, fill=blue!20]
    ]
    [Relation Extraction, fill=yellow!20
        [{
        CoT, PS, Self-Consistency, MP [\cite{wang2023metacognitive}]
        }, fill=yellow!20]
    ]
    [Natural Language Inference, fill=cyan!20
        [{
        CoT, PS, Self-Consistency, MP [\cite{wang2023metacognitive}]
        }, fill=cyan!20]
    ]
    [Stance Detection, fill=purple!20
        [{
        Basic, CoT [\cite{zhang2023investigating}]
        }, fill=purple!20]
    ]
    [Paraphrasing, fill=pink!20
        [{
        CoT, PS, Self-Consistency, MP [\cite{wang2023metacognitive}]
        }, fill=pink!20]
    ]
    [Summarization, fill=green!20
        [{
        CoE, Basic [\cite{bao2024chain}]
        }, fill=green!20]
    ]
    [Word Sense Disambiguation, fill=orange!20
        [{
        CoT, PS, Self-Consistency, MP [\cite{wang2023metacognitive}]
        }, fill=orange!20]
    ]
    [Named Entity Recognition, fill=blue!20
        [{
        Basic, Basic + Annotation Guideline-Based Prompting, \\
        Basic + Annotation Guideline-Based Prompting + \\
        Error Analysis-Based Prompting, CoT,
        PS, Self-Consistency, \\ MP [\cite{hu2024improving},
        \cite{tang2024large}, \cite{wang2023metacognitive}]
        }, fill=blue!20]
    ]
    [Machine Translation, fill=yellow!20
        [{
        Basic, CoT, CoC, Basic + Variations [\cite{li2023chain}, \\ \cite{zhang2023prompting}]
        }, fill=yellow!20]
    ]
    [Emotion/Sentiment \\ Understanding, fill=cyan!20
        [{
        Basic, CoT, CoC,  THOR, Basic + Variations [\cite{li2023chain}, \\ \cite{fei2023reasoning}, \cite{fatouros2023transforming}]
        }, fill=cyan!20]
    ]
    [Recommender System, fill=purple!20
        [{
        Basic, CoT, CoC [\cite{li2023chain}]
        }, fill=purple!20]
    ]
    [Table-Based Mathematical \\ Problem Solving, fill=pink!20
        [{
        PoT, CoT, Self-Consistency, PAL, Basic, CoC, Random CoT, \\ Complex CoT [\cite{chen2022program}, \cite{li2023chain}, \cite{gao2023pal}, \\ \cite{fu2022complexity}]
        }, fill=pink!20]
    ]
    [Table-Based \\ Question-Answering, fill=green!20
        [{
        Basic, CoT, Binder, Dater, Chain-of-Table, Self-Consistency, VE, 
        \\ CoK [\cite{wang2024chain}, \cite{li2023chaink},\cite{cheng2022binding}, \\
        \cite{ye2023large}]
        }, fill=green!20]
    ]    
]
\end{forest}
\caption{Taxonomy Diagram of Prompt Engineering Methods Across Different NLP Tasks}
\label{taxonomy}
\end{figure}

research papers. Another point to highlight is that in many works, the authors have experimented with different versions of the same dataset, making it difficult for an absolute comparison between different prompting techniques applied to them. Based on our understanding, we have considered all the above-mentioned factors and used our best judgment when selecting the SoTA for each dataset.

\normalsize
\subsection{Mathematical Problem Solving}
This task measures a model's ability to perform any kind of mathematical computation in a non tabular setting. The different datasets which we came across while reading up on different prompting methods for this task are GSM8K \cite{cobbe2021training}, MATH \cite{hendrycks2021measuring}, SVAMP \cite{patel2021nlp}, ASDiv \cite{miao2021diverse}, AQuA \cite{ling2017program}, MAWPS \cite{koncel2016mawps}, MultiArith \cite{koncel2016mawps}, AddSub \cite{koncel2016mawps}, SingleEq \cite{koncel2016mawps}, Game of 24 \cite{yao2024tree}, Multi-Step Arithmetic \cite{srivastava2022beyond}, GSM-HARD \cite{gao2023pal}, SingleOp \cite{koncel2016mawps} and MathQA \cite{amini2019mathqa}. Table \ref{tab:mps} lists above-mentioned datasets and different prompting methods that have been experimented on them along with the best performing prompting strategy.

\normalsize
\subsection{Logical Reasoning}

Logical Reasoning task checks a model's natural language understanding to follow a set of commands with inputs and solve a given problem. The different datasets which we covered while reading up on different prompting strategies for this task are Word Sorting \cite{srivastava2022beyond}, Temporal Sequences \cite{srivastava2022beyond}, Formal Fallacies \cite{srivastava2022beyond}, Mini Crosswords \cite{yao2024tree}, Object Counting \cite{srivastava2022beyond}, Logical Deduction \cite{srivastava2022beyond}, Boolean Expressions \cite{srivastava2022beyond}, Tracking Shuffled Objects \cite{srivastava2022beyond}, Web of Lies \cite{srivastava2022beyond}, Dyck Languages \cite{srivastava2022beyond}, Geometric Shapes \cite{srivastava2022beyond}, Repeat Copy Logic \cite{srivastava2022beyond}. Table \ref{tab:logicalr} contains above-mentioned datasets and different prompting techniques that have been experimented on them along with the best performing prompting method.

% \paragraph{Chain-Of-Thought (CoT)} 

% \begin{table*}[h]
%     \centering
%     \begin{tabular}{p{1.5in}p{1.5in}p{1.5in}p{1.5in}p{1.5in}}
%     \toprule
%         \textbf{Dataset}     & \textbf{ProcessBank} & \textbf{BioMRC}  & \textbf{MASH-QA} & \textbf{CliCR}   \\
%         \midrule
%         \# QA Pairs & 150     & 6250     & 3493 & 7184\\
%         Avg Context Length    & 85     & 255   & 863  & 1461 \\
%         Max Context Length    & 266    & 510  & 2911 & 3952\\
%         % MASH-QA     & 35k    & XX     &   696   \\
%     \end{tabular}
%     \caption{Corpus Level Statistics}
%     \label{tab:datset}
% \end{table*}

\footnotesize
\begin{longtable}{p{.6in}p{1.6in}p{2.6in}p{.5in}}
    \caption{Prompt Engineering Analysis for Mathematical Problem Solving Task}
    \label{tab:mps} \\
    \toprule
    \textbf{Dataset}    & \textbf{Prompting Strategies} & \textbf{LLM(s)}  & \textbf{SoTA}  \\
    \midrule
    \endfirsthead
    
    \multicolumn{4}{c}%
    % {{\bfseries \tablename\ \thetable{} -- Continued from the Previous Page}} \\
    {{ \tablename\ \thetable{} Continued from the Previous Page}} \\
    \midrule
    \textbf{Dataset}    & \textbf{Prompting Strategies} & \textbf{LLM(s)}  & \textbf{SoTA}  \\
    \midrule
    \endhead

    % \midrule 
    \multicolumn{4}{c}{{\tablename\ \thetable{} Continued on the Next Page}} \\
    \endfoot
    
    \bottomrule
    \endlastfoot

    % Table content
GSM8K   & Basic, Analogical Reasoning, CoT, Auto-CoT, Self-Consistency, LoT, PoT, PAL, CoC, Contrastive CoT, Contrastive Self-Consistency, Least-to-Most, Synthetic Prompting, Random CoT, Complex CoT, Active-Prompt, Fed-SP-SC, Fed-DP-CoT, PS    & GPT-3.5-Turbo, GPT-4, PaLM 2-L, GPT-3 (Text-Davinci-002), LaMDA-137B, PaLM-540B, UL2-20B, Codex (Code-Davinci-002), GPT-3, Codex (Code-Davinci-001), Vicuna-7B, Vicuna-13B, Vicuna-33B, CodeGen (Codegen-16B-Multi), CodeGen (Codegen-16B-Mono), CodeT5+, Xgen, PaLM, LaMDA, PaLM 2-S, GPT-3.5 (Text-Davinci-003), Minerva-540B, InstructGPT (Text-Davinci-003), DiVeRSe, UL2-20B   & PoT \\
        \midrule

        MATH   & Analogical Reasoning, CoT   & GPT-3.5-Turbo, GPT-4, PaLM 2-L    & Analogical Reasoning \\
        \midrule
        
        SVAMP   & Basic, CoT, Auto-CoT, Self-Consistency, PAL, PoT, Random CoT, Active-Prompt, Synthetic Prompting, Contrastive CoT, Contrastive Self-Consistency, Fed-SP-SC, Fed-DP-CoT, PS   & GPT-3 (Text-Davinci-002), LaMDA-137B, PaLM-540B, UL2-20B, Codex (Code-Davinci-002), GPT-3, UL2-20B, Codex (Code-Davinci-001), GPT-3.5-Turbo, CodeGen (Codegen-16B-Multi), CodeGen (Codegen-16B-Mono), CodeT5+, Xgen, PaLM, LaMDA, Minerva-540B, GPT-3.5 (Text-Davinci-003), InstructGPT (Text-Davinci-003)    & PoT \\
        \midrule
        % SVAMP   & Basic, CoT, Auto-CoT   & GPT-3.5 (Text-Davinci-002), Codex (Code-Davinci-002)    & 3493 \\
        % \midrule
        % SVAMP   & CoT, Self-Consistency   & GPT-3, LaMDA-137B, PaLM-540B, UL2-20B, Codex (Code-Davinci-001, Code-Davinci-002)    & 3493 \\
        % \midrule
        % SVAMP   & PoT, CoT, Self-Consistency, PAL  & Codex (Code-Davinci-002), GPT-3.5 (Text-Davinci-002), GPT-3.5-Turbo, CodeGen (Codegen-16B-Multi and Codegen-16B-Mono), CodeT5+, Xgen, PaLM, LaMDA   & 3493 \\
        % \midrule
        % SVAMP   & Basic, CoT, PAL & Codex (Code-Davinci-002), UL2-20B, LaMDA-137B, PaLM-540B, Minerva-540B   & 3493 \\
        % \midrule
        % SVAMP   & CoT, Fed-SP-SC, Fed-DP-SC, Fed-SP-CoT, Fed-DP-CoT & GPT-3.5 (Text-Davinci-002), GPT-3.5 (Text-Davinci-003)   & 3493 \\
        % \midrule
        % SVAMP   & Basic, CoT, Contrastive CoT, Contrastive Self-Consistency & GPT-3.5-Turbo   & 3493 \\
        % \midrule
        % SVAMP   & Basic, CoT, PAL, Synthetic Prompting & InstructGPT (Text-Davinci-003), Codex (Code-Davinci-002)  & 3493 \\
        % \midrule
        % SVAMP   & CoT, Auto-CoT, Self-Consistency, Random-CoT, Active-Prompt & UL2-20B, LaMDA-137B, PaLM-540B, GPT-3.5 (Text-Davinci-002), GPT-3.5 (Text-Davinci-003), Codex (Code-Davinci-002)  & 3493 \\
        % \midrule
        ASDiv   & Basic, CoT, Self-Consistency, PAL, Contrastive CoT, Contrastive Self-Consistency, Synthetic Prompting, Auto-CoT, Random CoT, Active-Prompt   & GPT-3 (Text-Davinci-002), LaMDA-137B, PaLM-540B, UL2-20B, Codex (Code-Davinci-002), GPT-3, Codex (Code-Davinci-001), Minerva-540B, GPT-3.5-Turbo, InstructGPT (Text-Davinci-003), GPT-3.5 (Text-Davinci-003)     & Contrastive Self-Consistency \\
        \midrule
        % ASDiv   & CoT, Self-Consistency   & GPT-3, LaMDA-137B, PaLM-540B, UL2-20B, Codex (Code-Davinci-001, Code-Davinci-002)    & 3493 \\
        % \midrule
        % ASDiv   & Basic, CoT, PAL & Codex (Code-Davinci-002), UL2-20B, LaMDA-137B, PaLM-540B, Minerva-540B   & 3493 \\
        % \midrule
        % ASDiv   & Basic, CoT, Contrastive CoT, Contrastive Self-Consistency & GPT-3.5-Turbo   & 3493 \\
        % \midrule
        % ASDiv   & Basic, CoT, PAL, Synthetic Prompting & InstructGPT (Text-Davinci-003), Codex (Code-Davinci-002)  & 3493 \\
        % \midrule
        % ASDiv   & CoT, Auto-CoT, Self-Consistency, Random-CoT, Active-Prompt & UL2-20B, LaMDA-137B, PaLM-540B, GPT-3.5 (Text-Davinci-002), GPT-3.5 (Text-Davinci-003), Codex (Code-Davinci-002)  & 3493 \\
        % \midrule
        AQuA   & Basic, CoT, Auto-CoT, Self-Consistency, LoT, PoT, Contrastive CoT, Contrastive Self-Consistency, Random CoT, Active-Prompt, PS   & GPT-3 (Text-Davinci-002), LaMDA-137B, PaLM-540B, UL2-20B, Codex (Code-Davinci-002), GPT-3, Codex (Code-Davinci-001), GPT-3.5-Turbo, GPT-4, Vicuna-7B, Vicuna-13B, Vicuna-33B, CodeGen (Codegen-16B-Multi), CodeGen (Codegen-16B-Mono), CodeT5+, Xgen, PaLM, LaMDA, GPT-3.5 (Text-Davinci-003)     & PoT \\
        \midrule
        % AQuA   & Basic, CoT, Auto-CoT   & GPT-3.5 (Text-Davinci-002), Codex (Code-Davinci-002)    & 3493 \\
        % \midrule
        % AQuA   & CoT, Self-Consistency   & GPT-3, LaMDA-137B, PaLM-540B, UL2-20B, Codex (Code-Davinci-001, Code-Davinci-002)    & 3493 \\
        % \midrule
        % AQuA   & CoT, LoT   & GPT-3.5-Turbo, GPT-4, Vicuna-7B, Vicuna-13B, Vicuna-33B   & 3493 \\
        % \midrule
        % AQuA   & PoT, CoT, Self-Consistency, PAL  & Codex (Code-Davinci-002), GPT-3.5 (Text-Davinci-002), GPT-3.5-Turbo, CodeGen (Codegen-16B-Multi and Codegen-16B-Mono), CodeT5+, Xgen, PaLM, LaMDA   & 3493 \\
        % \midrule
        % AQuA   & Basic, CoT, Contrastive CoT, Contrastive Self-Consistency & GPT-3.5-Turbo   & 3493 \\
        % \midrule
        % AQuA   & Basic, DSP & Codex (Code-Davinci-002), GPT-3.5-Turbo, InstructGPT (Text-Davinci-002)  & 3493 \\
        % \midrule
        % AQuA   & CoT, Auto-CoT, Self-Consistency, Random-CoT, Active-Prompt & UL2-20B, LaMDA-137B, PaLM-540B, GPT-3.5 (Text-Davinci-002), GPT-3.5 (Text-Davinci-003), Codex (Code-Davinci-002)  & 3493 \\
        % \midrule
        
        MAWPS   & Basic, CoT   & GPT-3 (Text-Davinci-002), LaMDA-137B, PaLM-540B, UL2-20B, Codex (Code-Davinci-002)    & CoT \\
        \midrule
        Game of 24   & Basic, CoT, Self-Consistency, ToT   & GPT-4    & ToT \\
        \midrule

        MultiArith   & Basic, CoT, Auto-CoT, Self-Consistency, PoT, PAL, MathPrompter, Random CoT, Complex CoT, PS   & GPT-3 (Text-Davinci-002), Codex (Code-Davinci-002), GPT-3, LaMDA-137B, PaLM-540B, UL2-20B, Codex (Code-Davinci-001), GPT-3.5-Turbo, CodeGen (Codegen-16B-Multi), CodeGen (Codegen-16B-Mono), CodeT5+, Xgen, PaLM, LaMDA, Minerva-540B, GPT-3.5 (Text-Davinci-003), DiVeRSe    & Self-Consistency \\
        \midrule

        Multi-Step Arithmetic   & Basic, CoT, CoC  & PaLM 2-S, GPT-3.5 (Text-Davinci-003), GPT-3.5-Turbo, GPT-4   & CoC \\
        \midrule
        % MultiArith   & CoT, Self-Consistency   & GPT-3, LaMDA-137B, PaLM-540B, UL2-20B, Codex (Code-Davinci-001, Code-Davinci-002)    & 3493 \\
        % \midrule
        % MultiArith   & PoT, CoT, Self-Consistency, PAL  & Codex (Code-Davinci-002), GPT-3.5 (Text-Davinci-002), GPT-3.5-Turbo, CodeGen (Codegen-16B-Multi and Codegen-16B-Mono), CodeT5+, Xgen, PaLM, LaMDA   & 3493 \\
        % \midrule
        % MultiArith   & Basic, CoT, PAL & Codex (Code-Davinci-002), UL2-20B, LaMDA-137B, PaLM-540B, Minerva-540B   & 3493 \\
        % \midrule

        % MultiArith   & Basic, CoT, Self-Consistency, MathPrompter & GPT-3.5 (Text-Davinci-003)  & 3493 \\
        % \midrule
        
        % MultiArith   & CoT, Random CoT, Complex CoT & LaMDA-137B, PaLM-540B, Minerva-540B, GPT-3.5 (Text-Davinci-002), Codex (Code-Davinci-002), DiVeRSe  & 3493 \\
        % \midrule
        AddSub   & Basic, CoT, Auto-CoT, Self-Consistency, PAL, PoT, PS    & GPT-3 (Text-Davinci-002), GPT-3.5 (Text-Davinci-003) Codex (Code-Davinci-002), UL2-20B, LaMDA-137B, PaLM-540B, Minerva-540B    & PAL \\
        \midrule
        % AddSub   & CoT, Self-Consistency   & GPT-3, LaMDA-137B, PaLM-540B, UL2-20B, Codex (Code-Davinci-001, Code-Davinci-002)    & 3493 \\
        % \midrule
        % AddSub   & Basic, CoT, PAL & Codex (Code-Davinci-002), UL2-20B, LaMDA-137B, PaLM-540B, Minerva-540B   & 3493 \\
        % \midrule
        
        SingleEq   & Basic, CoT, Auto-CoT, PAL, Self-Consistency, Random CoT, Active-Prompt, PS, PoT   & GPT-3 (Text-Davinci-002), Codex (Code-Davinci-002), UL2-20B, LaMDA-137B, PaLM-540B, Minerva-540B, GPT-3.5 (Text-Davinci-003)    & Active-Prompt \\
        \midrule
        % SingleEq   & Basic, CoT, PAL & Codex (Code-Davinci-002), UL2-20B, LaMDA-137B, PaLM-540B, Minerva-540B   & 3493 \\
        % \midrule
        % SingleEq   & CoT, Auto-CoT, Self-Consistency, Random-CoT, Active-Prompt & UL2-20B, LaMDA-137B, PaLM-540B, GPT-3.5 (Text-Davinci-002), GPT-3.5 (Text-Davinci-003), Codex (Code-Davinci-002)  & 3493 \\
        % \midrule

        GSM-HARD   & Basic, CoT, PAL, Contrastive CoT, Contrastive Self-Consistency, Synthetic Prompting & Codex (Code-Davinci-002), UL2-20B, LaMDA-137B, PaLM-540B, Minerva-540B, GPT-3.5-Turbo, InstructGPT (Text-Davinci-003)   & Synthetic Prompting \\
        \midrule

        % GSM-HARD   & Basic, CoT, Contrastive CoT, Contrastive Self-Consistency & GPT-3.5-Turbo   & 3493 \\
        % \midrule
    
        % GSM-HARD    & Basic, CoT, PAL, Synthetic Prompting & InstructGPT (Text-Davinci-003), Codex (Code-Davinci-002)  & 3493 \\
        % \midrule

        SingleOp   & Basic, CoT, PAL, Synthetic Prompting & Codex (Code-Davinci-002), UL2-20B, LaMDA-137B, PaLM-540B, Minerva-540B, InstructGPT (Text-Davinci-003), GPT-3 (Text-Davinci-002)   & Synthetic Prompting \\
        \midrule        
        
        % SingleOp   & Basic, CoT, PAL, Synthetic Prompting & InstructGPT (Text-Davinci-003), Codex (Code-Davinci-002)  & 3493 \\
        % \midrule

        MathQA   &  CoT, Random CoT, Complex CoT & LaMDA-137B, PaLM-540B, Minerva-540B, GPT-3 (Text-Davinci-002), Codex (Code-Davinci-002), DiVeRSe  & Complex CoT \\
        % \midrule
        % MultiArith   & Basic, DSP & Codex (Code-Davinci-002), GPT-3.5-Turbo, InstructGPT (Text-Davinci-002)  & 3493 \\
        % \midrule
\end{longtable}
\normalsize

% \begin{longtable}{p{.6in}p{1.6in}p{3.0in}p{.5in}}
%     \caption{Prompt Engineering Analysis for Mathematical Problem Solving Task}
%     \label{tab:mps} \\
%     \toprule
%     \textbf{Dataset}    & \textbf{Prompting Strategies} & \textbf{LLM(s)}  & \textbf{SoTA}  \\
%     \midrule
%     \endfirsthead
    
%     \multicolumn{4}{c}%
%     {{\bfseries \tablename\ \thetable{} -- continued from previous page}} \\
%     \toprule
%     \textbf{Dataset}    & \textbf{Prompting Strategies} & \textbf{LLM(s)}  & \textbf{SoTA}  \\
%     \midrule
%     \endhead

%     \midrule \multicolumn{4}{r}{{Continued on next page}} \\
%     \endfoot
    
%     \bottomrule
%     \endlastfoot

% \end{longtable}

\footnotesize
\begin{longtable}{p{.6in}p{1.6in}p{2.6in}p{.5in}}
    \caption{Prompt Engineering Analysis for Logical Reasoning Task}
    \label{tab:logicalr} \\
    \toprule
    \textbf{Dataset}    & \textbf{Prompting Strategies} & \textbf{LLM(s)}  & \textbf{SoTA}  \\
    \midrule
    \endfirsthead
    
    \multicolumn{4}{c}%
    {{ \tablename\ \thetable{} Continued from the Previous Page}} \\
    \midrule
    \textbf{Dataset}    & \textbf{Prompting Strategies} & \textbf{LLM(s)}  & \textbf{SoTA}  \\
    \midrule
    \endhead

    % \midrule 
    \multicolumn{4}{c}{{\tablename\ \thetable{} Continued on the Next Page}} \\
    \endfoot

    \bottomrule
    \endlastfoot

    Word Sorting   & Basic, Analogical Reasoning, CoT, CoC   & GPT-3.5-Turbo, GPT-4, PaLM 2-L, PaLM 2-S, GPT-3.5 (Text-Davinci-003)     & CoC \\
        \midrule
        % Word Sorting     & Basic, CoT, CoC  & PaLM 2-S, GPT-3.5 (Text-Davinci-003), GPT-3.5-Turbo, GPT-4     & 3493 \\
        % \midrule
        Logical Deduction   & Basic, Analogical Reasoning, CoT, CoC   & GPT-3.5-Turbo, GPT-4, PaLM 2-L, PaLM 2-S, GPT-3.5 (Text-Davinci-003)    & CoC \\
        \midrule
        % Logical Deduction     & Basic, CoT, CoC  & PaLM 2-S, GPT-3.5 (Text-Davinci-003), GPT-3.5-Turbo, GPT-4     & 3493 \\
        % \midrule
        Temporal Sequences   & Basic, Analogical Reasoning, CoT, CoC   & GPT-3.5-Turbo, GPT-4, PaLM 2-L, PaLM 2-S, GPT-3.5 (Text-Davinci-003)     & CoC \\
        \midrule
        % Temporal Sequences      & Basic, CoT, CoC  & PaLM 2-S, GPT-3.5 (Text-Davinci-003), GPT-3.5-Turbo, GPT-4     & 3493 \\
        % \midrule
        Formal Fallacies   & Basic, Analogical Reasoning, CoT, CoC   & GPT-3.5-Turbo, GPT-4, PaLM 2-L, PaLM 2-S, GPT-3.5 (Text-Davinci-003)     & Analogical Reasoning \\
        \midrule
        % Formal Fallacies     & Basic, CoT, CoC  & PaLM 2-S, GPT-3.5 (Text-Davinci-003), GPT-3.5-Turbo, GPT-4     & 3493 \\
        % \midrule
        Mini Crosswords    & Basic, CoT, ToT  & GPT-4     & ToT \\
        \midrule
        Tracking Shuffled Objects     & Basic, CoT, LoT, CoC  & GPT-3.5-Turbo, GPT-4, Vicuna-7B, Vicuna-13B, Vicuna-33B, PaLM 2-S, GPT-3.5 (Text-Davinci-003)     & CoT, LoT, CoC\\
        \midrule
        % Tracking Shuffled Objects     & Basic, CoT, CoC  & PaLM 2-S, GPT-3.5 (Text-Davinci-003), GPT-3.5-Turbo, GPT-4     & 3493 \\
        % \midrule
        Object Counting     & Basic, CoT, CoC, PAL  & PaLM 2-S, GPT-3.5 (Text-Davinci-003), GPT-3.5-Turbo, GPT-4, Codex (Code-Davinci-002), UL2-20B, LaMDA-137B, PaLM-540B, Minerva-540B     & CoC \\
        \midrule
        % Object Counting      & Basic, CoT, PAL  & Codex (Code-Davinci-002), UL2-20B, LaMDA-137B, PaLM-540B, Minerva-540B    & 3493 \\

        Boolean Expressions     & Basic, CoT, CoC  & PaLM 2-S, GPT-3.5 (Text-Davinci-003), GPT-3.5-Turbo, GPT-4     & CoC \\
        \midrule

        Web of Lies     & Basic, CoT, CoC  & PaLM 2-S, GPT-3.5 (Text-Davinci-003), GPT-3.5-Turbo, GPT-4     & CoT \\
        \midrule
        
        Dyck Languages     & Basic, CoT, CoC  & PaLM 2-S, GPT-3.5 (Text-Davinci-003), GPT-3.5-Turbo, GPT-4     & CoC \\
        \midrule
        
        Geometric Shapes      & Basic, CoT, CoC  & PaLM 2-S, GPT-3.5 (Text-Davinci-003), GPT-3.5-Turbo, GPT-4     & CoC \\
        \midrule
        Repeat Copy Logic       & Basic, CoT, PAL, Synthetic Prompting  & Codex (Code-Davinci-002), UL2-20B, LaMDA-137B, PaLM-540B, Minerva-540B, InstructGPT (Text-Davinci-003)    & PAL \\
        % \midrule
        % Repeat Copy Logic      & Basic, CoT, PAL, Synthetic Prompting  & InstructGPT (Text-Davinci-003), Codex (Code-Davinci-002)    & 3493 \\

\end{longtable}
\normalsize

\subsection{Commonsense Reasoning}

Contrary to Logical Reasoning task, Commonsense Reasoning task measures a model's ability in terms of common practical knowledge often referred to as \textit{commonsense} by humans to make any kind of judgement. It does not involve solving a problem to arrive at an answer. Rather, it is more of a form of inherent general knowledge. The various datasets that we discovered while surveying different prompting methods for this task include Reasoning about Colored Objects \cite{srivastava2022beyond}, CSQA \cite{talmor2018commonsenseqa}, Date Understanding \cite{srivastava2022beyond}, Sports Understanding \cite{srivastava2022beyond}, Last Letter Concatenation \cite{wei2022chain}, Coin Flip \cite{wei2022chain}, Odd One Out \cite{srivastava2022beyond}, Disambiguation QA \cite{srivastava2022beyond}, Hyperbaton \cite{srivastava2022beyond}, Com2Sense \cite{singh2021com2sense}, CSQA 2.0 \cite{talmor2022commonsenseqa}, Creak \cite{onoe2021creak} and List Reversal \cite{khot2022decomposed}. Table \ref{tab:commonr} shows above-mentioned datasets and different prompting strategies that have been experimented on them along with the best performing prompting method.

\footnotesize
\begin{longtable}{p{.6in}p{1.6in}p{2.6in}p{.5in}}
    \caption{Prompt Engineering Analysis for Commonsense Reasoning Task}
    \label{tab:commonr} \\
    \toprule
    \textbf{Dataset}    & \textbf{Prompting Strategies} & \textbf{LLM(s)}  & \textbf{SoTA}  \\
    \midrule
    \endfirsthead
    
    \multicolumn{4}{c}%
    {{ \tablename\ \thetable{} Continued from the Previous Page}} \\
    \midrule
    \textbf{Dataset}    & \textbf{Prompting Strategies} & \textbf{LLM(s)}  & \textbf{SoTA}  \\
    \midrule
    \endhead

    % \midrule 
    \multicolumn{4}{c}{{ \tablename\ \thetable{} Continued on the Next Page}} \\
    \endfoot
    
    \bottomrule
    \endlastfoot

    Reasoning about Colored Objects   & Analogical Reasoning, CoT, Basic, CoC, PAL, Synthetic Prompting   & PaLM 2-L, PaLM 2-S, GPT-3.5 (Text-Davinci-003), GPT-3.5-Turbo, GPT-4, UL2-20B, LaMDA-137B, PaLM-540B, Minerva-540B, InstructGPT (Text-Davinci-003), Codex (Code-Davinci-002)     & Synthetic Prompting \\
        \midrule
        CSQA  & Basic, CoT, Auto-CoT, Self-Consistency, Random CoT, Active-Prompt, PoT, PS    & Codex (Code-Davinci-001), Codex (Code-Davinci-002), GPT-3, GPT-3 (Text-Davinci-002), GPT-3.5 (Text-Davinci-003), LaMDA-137B, PaLM-540B, UL2-20B    & Active-Prompt \\
        \midrule
        Last Letter Concatenation  & Basic, CoT, Auto-CoT, Self-Consistency, LoT, Random CoT, Active-Prompt, Least-to-Most, DecomP, PS    & Codex (Code-Davinci-001), Codex (Code-Davinci-002), GPT-3, GPT-3 (Text-Davinci-002), GPT-3.5 (Text-Davinci-003), GPT-3.5-Turbo, GPT-4, InstructGPT (Text-Davinci-001), InstructGPT (Text-Davinci-002), LaMDA-137B, PaLM-540B, UL2-20B, Vicuna-13B, Vicuna-33B, Vicuna-7B     & DecomP \\
        \midrule

        CSQA 2.0   & Basic, CoT, Self-Consistency, GKP, Maieutic Prompting & InstructGPT (Text-Davinci-001)     & Maieutic Prompting \\
        \midrule
        
        Date Understanding  & Basic, CoT, LoT, CoC, PAL, Complex CoT    & Codex (Code-Davinci-002), DiVeRSe', GPT-3 (Text-Davinci-002), GPT-3.5 (Text-Davinci-003), GPT-3.5-Turbo, GPT-4, LaMDA-137B, Minerva-540B, PaLM 2-S, PaLM-540B, UL2-20B, Vicuna-13B, Vicuna-33B, Vicuna-7B & Complex CoT \\
        \midrule
        
        Sports Understanding  & Basic, CoT, CoC    & GPT-3 (Text-Davinci-002), LaMDA-137B, PaLM-540B, UL2-20B, Codex (Code-Davinci-002), PaLM 2-S, GPT-3.5 (Text-Davinci-003), GPT-3.5-Turbo, GPT-4     & CoT \\
        \midrule
        % SayCan  & Basic, CoT    & GPT-3.5 (Text-Davinci-002), LaMDA-137B, PaLM-540B, UL2-20B, Codex (Code-Davinci-002)     & 3493 \\
        % \midrule
        
        % Last Letter Concatenation  & Basic, CoT, Auto-CoT     & GPT-3.5 (Text-Davinci-002), Codex (Code-Davinci-002)     & 3493 \\
        % \midrule
        % Last Letter Concatenation  & CoT, Self-Consistency   & GPT-3, LaMDA-137B, PaLM-540B, UL2-20B, Codex (Code-Davinci-001, Code-Davinci-002)     & 3493 \\
        % \midrule
        % Last Letter Concatenation  & CoT, LoT   & GPT-3.5-Turbo, GPT-4, Vicuna-7B, Vicuna-13B, Vicuna-33B     & 3493 \\
        % \midrule
        % Last Letter Concatenation   & CoT, Auto-CoT, Self-Consistency, Random-CoT, Active-Prompt  & UL2-20B, LaMDA-137B, PaLM-540B, GPT-3.5 (Text-Davinci-002), GPT-3.5 (Text-Davinci-003), Codex (Code-Davinci-002)     & 3493 \\
        % \midrule
        % Last Letter Concatenation   & Basic, CoT, Least-to-Most & GPT-3.5 (Text-Davinci-002), Codex (Code-Davinci-002), Codex (Code-Davinci-001)     & 3493 \\
        % \midrule
        % Last Letter Concatenation    & CoT, DecomP, Least-to-Most & InstructGPT (Text-Davinci-002), InstructGPT (Text-Davinci-001), Codex (Code-Davinci-002)     & 3493 \\
        % \midrule
        Coin Flip  & Basic, CoT, Auto-CoT, Self-Consistency, PS   & GPT-3 (Text-Davinci-002), LaMDA-137B, PaLM-540B, UL2-20B, Codex (Code-Davinci-002), GPT-3, Codex (Code-Davinci-001)     & Auto-CoT \\
        \midrule
        % Coin Flip  & Basic, CoT, Auto-CoT    & GPT-3.5 (Text-Davinci-002), Codex (Code-Davinci-002)     & 3493 \\
        % \midrule
        % Coin Flip  & CoT, Self-Consistency   & GPT-3, LaMDA-137B, PaLM-540B, UL2-20B, Codex (Code-Davinci-001, Code-Davinci-002)     & 3493 \\
        % \midrule
        % CSQA  & Basic, CoT, Auto-CoT   & GPT-3.5 (Text-Davinci-002), Codex (Code-Davinci-002)     & 3493 \\
        % \midrule

        % CSQA  & CoT, Self-Consistency   & GPT-3, LaMDA-137B, PaLM-540B, UL2-20B, Codex (Code-Davinci-001, Code-Davinci-002)     & 3493 \\
        % \midrule
        % ARC  & CoT, Self-Consistency   & GPT-3, LaMDA-137B, PaLM-540B, UL2-20B, Codex (Code-Davinci-001, Code-Davinci-002)     & 3493 \\
        % \midrule

        % Date Understanding  & CoT, LoT   & GPT-3.5-Turbo, GPT-4, Vicuna-7B, Vicuna-13B, Vicuna-33B     & 3493 \\
        % \midrule
        Odd One Out  & CoT, LoT   & GPT-3.5-Turbo, GPT-4, Vicuna-7B, Vicuna-13B, Vicuna-33B     & LoT \\
        \midrule
        
        % Date Understanding   & Basic, CoT, CoC   & PaLM 2-S, GPT-3.5 (Text-Davinci-003), GPT-3.5-Turbo, GPT-4     & 3493 \\
        % \midrule
        Disambigu-ation QA    & Basic, CoT, CoC   & PaLM 2-S, GPT-3.5 (Text-Davinci-003), GPT-3.5-Turbo, GPT-4     & CoC \\
        \midrule
        % Sports Understanding   & Basic, CoT, CoC   & PaLM 2-S, GPT-3.5 (Text-Davinci-003), GPT-3.5-Turbo, GPT-4     & 3493 \\
        % \midrule
        Hyperbaton   & Basic, CoT, CoC   & PaLM 2-S, GPT-3.5 (Text-Davinci-003), GPT-3.5-Turbo, GPT-4     & CoC \\
        \midrule
        % Reasoning about Colored Objects   & Basic, CoT, CoC   & PaLM 2-S, GPT-3.5 (Text-Davinci-003), GPT-3.5-Turbo, GPT-4     & 3493 \\
        % \midrule
        % Date Understanding   & Basic, CoT, PAL   & Codex (Code-Davinci-002), UL2-20B, LaMDA-137B, PaLM-540B, Minerva-540B     & 3493 \\
        % \midrule
        % Reasoning about Colored Objects   & Basic, CoT, PAL   & Codex (Code-Davinci-002), UL2-20B, LaMDA-137B, PaLM-540B, Minerva-540B     & 3493 \\
        % \midrule
        % CSQA   & CoT, Auto-CoT, Self-Consistency, Random-CoT, Active-Prompt  & UL2-20B, LaMDA-137B, PaLM-540B, GPT-3.5 (Text-Davinci-002), GPT-3.5 (Text-Davinci-003), Codex (Code-Davinci-002)     & 3493 \\
        % \midrule
        
        % Reasoning about Colored Objects   & Basic, CoT, PAL, Synthetic Prompting  & InstructGPT (Text-Davinci-003), Codex (Code-Davinci-002)     & 3493 \\
        % \midrule
        Com2Sense   & Basic, CoT, Self-Consistency, GKP, Maieutic Prompting & InstructGPT (Text-Davinci-001)     & Maieutic Prompting \\
        \midrule

        Creak   & Basic, CoT, Self-Consistency, GKP, Maieutic Prompting & InstructGPT (Text-Davinci-001)     & Maieutic Prompting \\
        \midrule
        
        % Date Understanding   &  CoT, Random CoT, Complex CoT & LaMDA-137B, PaLM-540B, Minerva-540B, GPT-3.5 (Text-Davinci-002), Codex (Code-Davinci-002), DiVeRSe     & 3493 \\
        % \midrule
        
        List Reversal    & CoT, DecomP & InstructGPT (Text-Davinci-002), InstructGPT (Text-Davinci-001), Codex (Code-Davinci-002)     & DecomP \\
    
\end{longtable}

\normalsize

\subsection{Multi-Hop Reasoning}

Multi-Hop Reasoning task assess how good a model is at connecting pieces of evidence from different parts of a context to answer a given query. The different datasets which we covered while reading up on different prompting strategies for this task are StrategyQA \cite{geva2021did}, HotpotQA \cite{yang2018hotpotqa}, Bamboogle \cite{press2022measuring}, CommaQA-E \cite{khot2021hey}, MuSiQue \cite{trivedi2022musique}, 2WikiMultihopQA and \cite{ho2020constructing}. Table \ref{tab:mhr} lists above-mentioned datasets and different prompting methods that have been experimented on them along with the best performing prompting strategy.

\normalsize

\subsection{Causal Reasoning}

Causal Reasoning task checks a model's ability to deal with cause and effect. We came across two datasets while reading up on different prompting techniques for this task which are Cause And Effect \cite{srivastava2022beyond} and Causal Judgement \cite{srivastava2022beyond}. Table \ref{tab:causalr} shows above-mentioned datasets and different prompting techniques that have been experimented on them along with the best performing prompting method.

\footnotesize
\begin{longtable}{p{.6in}p{1.6in}p{2.6in}p{.5in}}
    \caption{Prompt Engineering Analysis for Multi-Hop Reasoning Task}
    \label{tab:mhr} \\
    \toprule
    \textbf{Dataset}    & \textbf{Prompting Strategies} & \textbf{LLM(s)}  & \textbf{SoTA}  \\
    \midrule
    \endfirsthead
    
    \multicolumn{4}{c}%
    {{ \tablename\ \thetable{} Continued from the Previous Page}} \\
    \midrule
    \textbf{Dataset}    & \textbf{Prompting Strategies} & \textbf{LLM(s)}  & \textbf{SoTA}  \\
    \midrule
    \endhead

    \multicolumn{4}{c}{{\tablename\ \thetable{} Continued on the Next Page}} \\
    \endfoot
    
    \bottomrule
    \endlastfoot

    StrategyQA   & Basic, CoT, Auto-CoT, Self-Consistency, Contrastive CoT, Contrastive Self-Consistency, Random CoT, Active-Prompt, Complex CoT, PS    & GPT-3, GPT-3 (Text-Davinci-002), GPT-3.5 (Text-Davinci-003), LaMDA-137B, PaLM-540B, UL2-20B, Codex (Code-Davinci-002), Codex (Code-Davinci-001), GPT-3.5-Turbo, Minerva-540B, DiVeRSe     & Active-Prompt \\
        \midrule
        % StrategyQA  & Basic, CoT, Auto-CoT   & GPT-3.5 (Text-Davinci-002), Codex (Code-Davinci-002)     & 3493 \\
        % \midrule
        % StrategyQA  & CoT, Self-Consistency & GPT-3, LaMDA-137B, PaLM-540B, UL2-20B, Codex (Code-Davinci-001, Code-Davinci-002)   & 3493 \\
        % \midrule
        HotpotQA  & Basic, CoT, Act, ReAct, Self-Consistency, VE, CoK, DecomP, Least-to-Most  & PaLM-540B, GPT-3 (Text-Davinci-002), GPT-3.5-Turbo, InstructGPT (Text-Davinci-002), InstructGPT (Text-Davinci-001), Codex (Code-Davinci-002)     & CoK \\
        \midrule
        % HotpotQA  & Basic, CoT, Self-Consistency, VE, CoK    & GPT-3.5-Turbo     & 3493 \\
        % \midrule
        % StrategyQA  & Basic, CoT, Contrastive CoT, Contrastive Self-Consistency    & GPT-3.5-Turbo     & 3493 \\
        % \midrule
        % StrategyQA  & CoT, Auto-CoT, Self-Consistency, Random-CoT, Active-Prompt   & UL2-20B, LaMDA-137B, PaLM-540B, GPT-3.5 (Text-Davinci-002), GPT-3.5 (Text-Davinci-003), Codex (Code-Davinci-002)     & 3493 \\
        % \midrule
        % StrategyQA  &  CoT, Random CoT, Complex CoT   & LaMDA-137B, PaLM-540B, Minerva-540B, GPT-3.5 (Text-Davinci-002), Codex (Code-Davinci-002), DiVeRSe    & 3493 \\
        % \midrule
        CommaQA-E  & CoT, DecomP   & InstructGPT (Text-Davinci-002), InstructGPT (Text-Davinci-001), Codex (Code-Davinci-002)    & DecomP \\
        % \midrule
        % HotpotQA  & CoT, DecomP, Least-to-Most   & InstructGPT (Text-Davinci-002), InstructGPT (Text-Davinci-001), Codex (Code-Davinci-002)    & 3493 \\
        \midrule
        MuSiQue  & Basic, CoT, DecomP   & InstructGPT (Text-Davinci-002), InstructGPT (Text-Davinci-001), Codex (Code-Davinci-002)   & DecomP \\
        \midrule
        2WikiMult-ihopQA  & Basic, CoT, DecomP   & InstructGPT (Text-Davinci-002), InstructGPT (Text-Davinci-001), Codex (Code-Davinci-002)    & DecomP \\
    
\end{longtable}

% \begin{longtable}{p{.6in}p{1.6in}p{3.0in}p{.5in}}
%     \caption{Prompt Engineering Analysis for Mathematical Problem Solving Task}
%     \label{tab:mps} \\
%     \toprule
%     \textbf{Dataset}    & \textbf{Prompting Strategies} & \textbf{LLM(s)}  & \textbf{SoTA}  \\
%     \midrule
%     \endfirsthead
    
%     \multicolumn{4}{c}%
%     {{\bfseries \tablename\ \thetable{} -- continued from previous page}} \\
%     \toprule
%     \textbf{Dataset}    & \textbf{Prompting Strategies} & \textbf{LLM(s)}  & \textbf{SoTA}  \\
%     \midrule
%     \endhead

%     \midrule \multicolumn{4}{r}{{Continued on next page}} \\
%     \endfoot
    
%     \bottomrule
%     \endlastfoot

% \end{longtable}

\footnotesize
\begin{longtable}{p{.6in}p{1.6in}p{2.6in}p{.5in}}
    \caption{Prompt Engineering Analysis for Causal Reasoning Task}
    \label{tab:causalr} \\
    \toprule
    \textbf{Dataset}    & \textbf{Prompting Strategies} & \textbf{LLM(s)}  & \textbf{SoTA}  \\
    \midrule
    \endfirsthead
    
    \multicolumn{4}{c}%
    {{\bfseries \tablename\ \thetable{} -- continued from previous page}} \\
    \toprule
    \textbf{Dataset}    & \textbf{Prompting Strategies} & \textbf{LLM(s)}  & \textbf{SoTA}  \\
    \midrule
    \endhead

    \midrule \multicolumn{4}{r}{{Continued on next page}} \\
    \endfoot
    
    \bottomrule
    \endlastfoot

    Cause And Effect   & CoT, LoT    & GPT-3.5-Turbo, GPT-4, Vicuna-7B, Vicuna-13B, Vicuna-33B      & LoT \\
        \midrule
        Causal Judgement  & Basic, CoT, CoC    & PaLM 2-S, GPT-3.5 (Text-Davinci-003), GPT-3.5-Turbo, GPT-4     & Basic, CoT \\
    
\end{longtable}
\normalsize

\subsection{Social Reasoning}

This task tests a model's ability to reason about human social interactions. We discovered only one dataset while surveying different prompting techniques for this task which is SocialQA \cite{srivastava2022beyond}. Table \ref{tab:socialr} contains above-mentioned datasets and different prompting methods that have been experimented on them along with the best performing prompting strategy.

\normalsize

\subsection{Contextual Question-Answering}

This task measures a model's ability to answer a query solely by relying on a given context. The different datasets which we covered while reading up on different prompting methods for this task are ProcessBank \cite{berant2014modeling}, BioMRC \cite{pappas2020biomrc}, MASH-QA \cite{zhu2020question}, CliCR \cite{vsuster2018clicr}, MultiSpanQA \cite{li2022multispanqa}, FinQA \cite{chen2021finqa}, TAT-QA \cite{zhu2021tat}, Patient Case Reports \cite{vatsal2024can}, Drop \cite{dua2019drop} and BoolQ \cite{clark2019boolq}. Table \ref{tab:cqa} lists above-mentioned datasets and different prompting methods that have been experimented on them along with the best performing prompting technique.

\footnotesize
\begin{longtable}{p{.6in}p{1.6in}p{2.6in}p{.5in}}
    \caption{Prompt Engineering Analysis for Social Reasoning Task}
    \label{tab:socialr} \\
    \toprule
    \textbf{Dataset}    & \textbf{Prompting Strategies} & \textbf{LLM(s)}  & \textbf{SoTA}  \\
    \midrule
    \endfirsthead
    
    \multicolumn{4}{c}%
    {{\bfseries \tablename\ \thetable{} -- continued from previous page}} \\
    \toprule
    \textbf{Dataset}    & \textbf{Prompting Strategies} & \textbf{LLM(s)}  & \textbf{SoTA}  \\
    \midrule
    \endhead

    \midrule \multicolumn{4}{r}{{Continued on next page}} \\
    \endfoot
    
    \bottomrule
    \endlastfoot

    SocialQA  & CoT, LoT    & GPT-3.5-Turbo, GPT-4, Vicuna-7B, Vicuna-13B, Vicuna-33B      & LoT \\
    
\end{longtable}

\footnotesize
\begin{longtable}{p{.6in}p{1.6in}p{2.6in}p{.5in}}
    \caption{Prompt Engineering Analysis for Contextual Question-Answering Task}
    \label{tab:cqa} \\
    \toprule
    \textbf{Dataset}    & \textbf{Prompting Strategies} & \textbf{LLM(s)}  & \textbf{SoTA}  \\
    \midrule
    \endfirsthead
    
    \multicolumn{4}{c}%
    {{ \tablename\ \thetable{} Continued from the Previous Page}} \\
    \toprule
    \textbf{Dataset}    & \textbf{Prompting Strategies} & \textbf{LLM(s)}  & \textbf{SoTA}  \\
    \midrule
    \endhead

    % \midrule 
    \multicolumn{4}{c}{{ \tablename\ \thetable{} Continued on the Next Page}} \\
    \endfoot
    
    \bottomrule
    \endlastfoot

    ProcessBank   & Basic, Implicit RAG, CoT, Analogical Reasoning    & GPT-4    & Implicit RAG \\
        \midrule
        BioMRC   & Basic, Implicit RAG, CoT, Analogical Reasoning   & GPT-4    & Basic \\
        \midrule
        MASH-QA   & Basic, Implicit RAG, CoT, Analogical Reasoning   & GPT-4    & Basic \\
        \midrule
        CliCR   & Basic, Implicit RAG, CoT, Analogical Reasoning   & GPT-4    & Implicit RAG, Analogical Reasoning  \\
        \midrule
        MultiSpanQA   & Basic, CoT, CoVe    & LLaMA-65B, LLaMA-2-70B Chat    & CoVe \\
        \midrule
        FinQA   & PoT, CoT, Self-Consistency   & Codex (Code-Davinci-002), GPT-3 (Text-Davinci-002), GPT-3.5-Turbo, CodeGen (Codegen-16B-Multi and Codegen-16B-Mono), CodeT5+, Xgen, PaLM, LaMDA    & PoT \\
        \midrule
        TAT-QA    & PoT, CoT, Self-Consistency   & Codex (Code-Davinci-002), GPT-3 (Text-Davinci-002), GPT-3.5-Turbo, CodeGen (Codegen-16B-Multi and Codegen-16B-Mono), CodeT5+, Xgen, PaLM, LaMDA    & PoT \\
        \midrule
        Patient Case Reports    & Implicit RAG, CoT, Analogical Reasoning, Basic, Basic with Term Definitions   & GPT-4    & Implicit RAG \\
        \midrule
        Drop    & Basic, CoT, Least-to-Most   & GPT-3 (Text-Davinci-002), Codex (Code-Davinci-002), Codex (Code-Davinci-001)   & Least-to-Most \\
        \midrule
        BoolQ    & CoT, PS, Self-Consistency, MP  & Llama-2-13B-Chat, GPT-3.5-Turbo, GPT-4, PaLM-Bison-Chat   & MP \\
    
\end{longtable}

\normalsize

\subsection{Context-Free Question-Answering}

In contrast to the Contextual Question-Answering task, the Context-Free Question-Answering task relies on model's embedded knowledge base or any open-source knowledge base, such as Wikipedia, to answer a query instead of using only the context provided.  The various datasets that we discovered while surveying different prompting techniques for this task are PopQA \cite{mallen2022not}, EntityQ \cite{sciavolino2021simple}, Wikidata \cite{dhuliawala2023chain}, Wiki-Catoegory List \cite{dhuliawala2023chain}, MedMCQA \cite{pal2022medmcqa}, MMLU Physics \cite{hendrycks2020measuring}, MMLU Biology  \cite{hendrycks2020measuring}, USMLE Sample Exam \cite{nori2023capabilities}, USMLE Self Assessments \cite{nori2023capabilities}, MedQA \cite{jin2021disease}, PubMedQA \cite{jin2019pubmedqa}, MMLU \cite{hendrycks2020measuring} and AI2 Reasoning Challenge \cite{clark2018think}. Table \ref{tab:cfqa} lists above-mentioned datasets and different prompting strategies that have been experimented on them along with the best performing prompting strategy.

\footnotesize
\begin{longtable}{p{.6in}p{1.6in}p{2.6in}p{.5in}}
    \caption{Prompt Engineering Analysis for Context-Free Question-Answering Task}
    \label{tab:cfqa} \\
    \toprule
    \textbf{Dataset}    & \textbf{Prompting Strategies} & \textbf{LLM(s)}  & \textbf{SoTA}  \\
    \midrule
    \endfirsthead
    
    \multicolumn{4}{c}%
    {{ \tablename\ \thetable{} Continued from the Previous Page}} \\
    \midrule
    \textbf{Dataset}    & \textbf{Prompting Strategies} & \textbf{LLM(s)}  & \textbf{SoTA}  \\
    \midrule
    \endhead

    % \midrule 
    \multicolumn{4}{c}{{ \tablename\ \thetable{} Continued on the Next Page}} \\
    \endfoot
    
    \bottomrule
    \endlastfoot

    PopQA  & Basic, CoT, ThoT    & GPT-4, GPT-3.5-Turbo, LLaMA-2-7B-Chat, LLaMA-2-13B-Chat, LLaMA-2-70B-Chat, Vicuna-7B, Vicuna-13B, Vicuna-33B      & ThoT \\
        \midrule
        EntityQ  & Basic, CoT, ThoT    & GPT-4, GPT-3.5-Turbo, LLaMA-2-7B-Chat, LLaMA-2-13B-Chat, LLaMA-2-70B-Chat, Vicuna-7B, Vicuna-13B, Vicuna-33B      & ThoT \\
        \midrule
        Wikidata  & Basic, CoT, CoVe     & LLaMA-65B, LLaMA-2-70B Chat      & CoVe \\
        \midrule
        Wiki-Catoegory List  & Basic, CoT, CoVe    & LLaMA-65B, LLaMA-2-70B Chat      & CoVe \\
        \midrule
        MedMCQA  & Basic, CoT, Self-Consistency, VE, CoK, ER    & GPT-3.5-Turbo, GPT-4, GPT-3.5, InstructGPT (Text-Davinci-002), Flan-PaLM 540B, Med-PaLM, Med-PaLM 2, Flan-PaLM, GPT-4-Base, Codex (Code-Davinci-002), LLaMA-2-70B, LLaMA-2-7B, LLaMA-2-13B, LLaMA-2-70B Chat, LLaMA-2-7B Chat, LLaMA-2-13B Chat, GPT-NeoX, MPT-Instruct-7B, MPT-Instruct-30B, Falcon-Instruct-7B, Falcon-Instruct-40B, Guanaco-33B, Guanaco-65B, Vicuna-1.3-7B, Vicuna-1.3-13B, Vicuna-1.3-33B, Vicuna-1.5-7B, Vicuna-1.5-13B, U-PaLM-540B, Flan-U-PaLM-540B, Med-PaLM V2-540B       & Basic \\
        \midrule

        MedQA  & Basic, CoT, Self-Consistency, ER    & GPT-4, GPT-3.5, GPT-3.5-Turbo, InstructGPT (Text-Davinci-002), Flan-PaLM 540B, Med-PaLM, Med-PaLM 2, Flan-PaLM, GPT-4-Base, Codex (Code-Davinci-002), LLaMA-2-70B, LLaMA-2-7B, LLaMA-2-13B, LLaMA-2-70B Chat, LLaMA-2-7B Chat, LLaMA-2-13B Chat, GPT-NeoX, MPT-Instruct-7B, MPT-Instruct-30B, Falcon-Instruct-7B, Falcon-Instruct-40B, Guanaco-33B, Guanaco-65B, Vicuna-1.3-7B, Vicuna-1.3-13B, Vicuna-1.3-33B, Vicuna-1.5-7B, Vicuna-1.5-13B, U-PaLM-540B, Flan-U-PaLM-540B, Med-PaLM V2-540B       & Basic \\
        \midrule
        
        MMLU Physics  & Basic, CoT, Self-Consistency, VE, CoK    & GPT-3.5-Turbo      & CoK \\
        \midrule
        MMLU Biology  & Basic, CoT, Self-Consistency, VE, CoK    & GPT-3.5-Turbo      & CoK \\
        \midrule
        USMLE Sample Exam & Basic    & GPT-4, GPT-3.5, GPT-3.5-Turbo, InstructGPT (Text-Davinci-002), Flan-PaLM 540B, Med-PaLM      & Basic \\
        \midrule
        USMLE Self Assessments  & Basic    & GPT-4, GPT-3.5, GPT-3.5-Turbo, InstructGPT (Text-Davinci-002), Flan-PaLM 540B, Med-PaLM      & Basic \\
        \midrule

        AI2 Reasoning Challenge  & CoT, Self-Consistency    & GPT-3, LaMDA-137B, PaLM-540B, UL2-20B, Codex (Code-Davinci-001), Codex (Code-Davinci-002)      & Self-Consistency \\
        \midrule

        % MedQA-USMLE  & Basic, CoT    & InstructGPT (Text-Davinci-002), Codex (Code-Davinci-002), LLaMA-2-70B, LLaMA-2-7B, LLaMA-2-13B, LLaMA-2-70B Chat, LLaMA-2-7B Chat, LLaMA-2-13B Chat, GPT-4, GPT-NeoX, MPT-Instruct-7B, MPT-Instruct-30B, Falcon-Instruct-7B, Falcon-Instruct-40B, Guanaco-33B, Guanaco-65B, Vicuna-1.3-7B, Vicuna-1.3-13B, Vicuna-1.3-33B, Vicuna-1.5-7B, Vicuna-1.5-13B, U-PaLM-540B, Flan-U-PaLM-540B, Med-PaLM V2-540B       & 3493 \\
        % \midrule
        PubMedQA  & Basic, CoT, Self-Consistency, ER    & GPT-4, GPT-3.5, GPT-3.5-Turbo, InstructGPT (Text-Davinci-002), Flan-PaLM 540B, Med-PaLM, Med-PaLM 2, Flan-PaLM, GPT-4-Base, Codex (Code-Davinci-002), LLaMA-2-70B, LLaMA-2-7B, LLaMA-2-13B, LLaMA-2-70B Chat, LLaMA-2-7B Chat, LLaMA-2-13B Chat, GPT-NeoX, MPT-Instruct-7B, MPT-Instruct-30B, Falcon-Instruct-7B, Falcon-Instruct-40B, Guanaco-33B, Guanaco-65B, Vicuna-1.3-7B, Vicuna-1.3-13B, Vicuna-1.3-33B, Vicuna-1.5-7B, Vicuna-1.5-13B, U-PaLM-540B, Flan-U-PaLM-540B, Med-PaLM V2-540B      & Basic \\
        \midrule
        % MedMCQA  & Basic    & GPT-4, GPT-3.5, GPT-3.5-Turbo, InstructGPT (Text-Davinci-002), Flan-PaLM 540B, Med-PaLM      & 3493 \\
        % \midrule
        % MMLU  & Basic    & GPT-4, GPT-3.5, GPT-3.5-Turbo, InstructGPT (Text-Davinci-002), Flan-PaLM 540B, Med-PaLM      & 3493 \\
        % \midrule
        % PubMedQA  & Basic, CoT, Self-Consistency, ER    & Med-PaLM 2, GPT-4, Flan-PaLM, GPT-4-Base      & 3493 \\
        % \midrule
        % MedMCQA  & Basic, CoT, Self-Consistency, ER    & Med-PaLM 2, GPT-4, Flan-PaLM, GPT-4-Base      & 3493 \\
        % \midrule
        MMLU  & Basic, CoT, Self-Consistency, ER    & Med-PaLM 2, Flan-PaLM, GPT-4-Base, GPT-4, GPT-3.5, GPT-3.5-Turbo, InstructGPT (Text-Davinci-002), Flan-PaLM 540B, Med-PaLM, Codex (Code-Davinci-002), LLaMA-2-70B, LLaMA-2-7B, LLaMA-2-13B, LLaMA-2-70B Chat, LLaMA-2-7B Chat, LLaMA-2-13B Chat, GPT-4, GPT-NeoX, MPT-Instruct-7B, MPT-Instruct-30B, Falcon-Instruct-7B, Falcon-Instruct-40B, Guanaco-33B, Guanaco-65B, Vicuna-1.3-7B, Vicuna-1.3-13B, Vicuna-1.3-33B, Vicuna-1.5-7B, Vicuna-1.5-13B, U-PaLM-540B, Flan-U-PaLM-540B, Med-PaLM V2-540B      & Basic \\

\end{longtable}
\normalsize

\subsection{Spatial Question-Answering}

Spatial Question-Answering task measures a model's ability to deal with spatial reasoning which is a cognitive process based on the construction of mental representations for spatial objects, relations, and transformations. The various datasets which we came across while reading up on different prompting techniques for this task include Brick World \cite{hu2023chain}, NLVR-Based Manipulation \cite{hu2023chain}, Natural Language Navigation \cite{hu2023chain}, Spartun \cite{mirzaee2022transfer} and Navigate \cite{srivastava2022beyond}. Table \ref{tab:sqa} contains above-mentioned datasets and different prompting methods that have been experimented on them along with the best performing prompting strategy.

\normalsize
\subsection{Conversational Contextual Question-Answering}

In this task, the model is assessed based on it's understanding of a given text extract and how it is able to answer a series of interconnected queries that appear in a conversational format. A key thing to note here is that each query may depend on previous queries' answers. We covered only one dataset while reading up on different prompting methods for this task which includes ConvFinQA \cite{chen2022convfinqa}. Table \ref{tab:ccqa} shows above-mentioned datasets and different prompting methods that have been experimented on them along with the best performing prompting strategy.

\footnotesize
\begin{longtable}{p{.6in}p{1.6in}p{2.6in}p{.5in}}
    \caption{Prompt Engineering Analysis for Spatial Question-Answering Task}
    \label{tab:sqa} \\
    \toprule
    \textbf{Dataset}    & \textbf{Prompting Strategies} & \textbf{LLM(s)}  & \textbf{SoTA}  \\
    \midrule
    \endfirsthead
    
    \multicolumn{4}{c}%
    {{ \tablename\ \thetable{} Continued from the Previous Page}} \\
    \midrule
    \textbf{Dataset}    & \textbf{Prompting Strategies} & \textbf{LLM(s)}  & \textbf{SoTA}  \\
    \midrule
    \endhead

    % \midrule 
    \multicolumn{4}{c}{{\tablename\ \thetable{} Continued on the Next Page}} \\
    \endfoot
    
    \bottomrule
    \endlastfoot

    Brick World   & CoT, CoS   & GPT-3.5 (Text-Davinci-003), GPT-3.5-Turbo, GPT-4     & CoS \\
        \midrule
        NLVR-Based Manipulation  & CoT, CoS    & GPT-3.5 (Text-Davinci-003), GPT-3.5-Turbo, GPT-4     & CoS \\
        \midrule
        Natural Language Navigation  & CoT, CoS   & GPT-3.5 (Text-Davinci-003), GPT-3.5-Turbo, GPT-4    & CoS \\
        \midrule
        Spartun  & CoT, CoS    & GPT-3.5 (Text-Davinci-003), GPT-3.5-Turbo, GPT-4     & CoS \\
        \midrule
        Navigate  & Basic, CoT, CoC    & PaLM 2-S, GPT-3.5 (Text-Davinci-003), GPT-3.5-Turbo, GPT-4     & CoT \\
    
\end{longtable}
\normalsize

\footnotesize
\begin{longtable}{p{.6in}p{1.6in}p{2.6in}p{.5in}}
    \caption{Prompt Engineering Analysis for Conversational Contextual Question-Answering Task}
    \label{tab:ccqa} \\
    \toprule
    \textbf{Dataset}    & \textbf{Prompting Strategies} & \textbf{LLM(s)}  & \textbf{SoTA}  \\
    \midrule
    \endfirsthead
    
    \multicolumn{4}{c}%
    {{\bfseries \tablename\ \thetable{} -- continued from previous page}} \\
    \toprule
    \textbf{Dataset}    & \textbf{Prompting Strategies} & \textbf{LLM(s)}  & \textbf{SoTA}  \\
    \midrule
    \endhead

    \midrule \multicolumn{4}{r}{{Continued on next page}} \\
    \endfoot
    
    \bottomrule
    \endlastfoot

    ConvFinQA  & PoT, CoT, Self-Consistency, PAL    & Codex (Code-Davinci-002), GPT-3 (Text-Davinci-002), GPT-3.5-Turbo, CodeGen (Codegen-16B-Multi), CodeGen (Codegen-16B-Mono), CodeT5+, Xgen, PaLM, LaMDA     & PoT \\
    
\end{longtable}

\subsection{Dialogue System}

Dialogue System task checks model's ability to perform language generation in a user-to-machine conversation setting or answer queries given an already generated conversation. It is possible that when the text extract in case of Conversational Contextual Question-Answering becomes a conversation, there will be a strong overlap between these two tasks but based on the datasets and prompting techniques encountered during our survey, we decided to keep these two as separate tasks. We discovered only one dataset while surveying different prompting methods for this task which includes Multi-Turn Conversation Response (MTCR) \cite{zhou2023thread}. Table \ref{tab:dias} lists above-mentioned datasets and different prompting strategies that have been experimented on them along with the best performing prompting technique.

\normalsize

\subsection{Code Generation}

This task involves all the cases where the input or the final output to the model is a programming language code. The different datasets which we came across while reading up on different prompting strategies for this task are Codeforce Scraping \cite{yasunaga2023large}, HumanEval \cite{chen2021evaluating}, MBPP \cite{austin2021program} and MBCPP \cite{athiwaratkun2022multi}. Table \ref{tab:codeg} contains above-mentioned datasets and different prompting techniques that have been experimented on them along with the best performing prompting strategy.

\footnotesize
\begin{longtable}{p{.6in}p{1.6in}p{2.6in}p{.5in}}
    \caption{Prompt Engineering Analysis for Dialogue System Task}
    \label{tab:dias} \\
    \toprule
    \textbf{Dataset}    & \textbf{Prompting Strategies} & \textbf{LLM(s)}  & \textbf{SoTA}  \\
    \midrule
    \endfirsthead
    
    \multicolumn{4}{c}%
    {{\bfseries \tablename\ \thetable{} -- continued from previous page}} \\
    \toprule
    \textbf{Dataset}    & \textbf{Prompting Strategies} & \textbf{LLM(s)}  & \textbf{SoTA}  \\
    \midrule
    \endhead

    \midrule \multicolumn{4}{r}{{Continued on next page}} \\
    \endfoot
    
    \bottomrule
    \endlastfoot

    MTCR  & Basic, CoT, ThoT    & GPT-4, GPT-3.5-Turbo, LLaMA-2-7B-Chat, LLaMA-2-13B-Chat, LLaMA-2-70B-Chat, Vicuna-7B, Vicuna-13B, Vicuna-33B      & ThoT \\
    
\end{longtable}

\footnotesize
\begin{longtable}{p{.6in}p{1.6in}p{2.6in}p{.5in}}
    \caption{Prompt Engineering Analysis for Code Generation Task}
    \label{tab:codeg} \\
    \toprule
    \textbf{Dataset}    & \textbf{Prompting Strategies} & \textbf{LLM(s)}  & \textbf{SoTA}  \\
    \midrule
    \endfirsthead
    
    \multicolumn{4}{c}%
    {{ \tablename\ \thetable{} Continued from the Previous Page}} \\
    \toprule
    \textbf{Dataset}    & \textbf{Prompting Strategies} & \textbf{LLM(s)}  & \textbf{SoTA}  \\
    \midrule
    \endhead

    % \midrule 
    \multicolumn{4}{c}{{\tablename\ \thetable{} Continued on the Next Page}} \\
    \endfoot
    
    \bottomrule
    \endlastfoot

    Codeforce Scraping   & Analogical Reasoning, CoT   & GPT-3.5-Turbo, GPT-4, PaLM 2-L    & Analogical Reasoning \\
        \midrule
        HumanEval  & Basic, SCoT, CoT   & Codex (Code-Davinci-002), GPT-3.5-Turbo     & SCoT \\
        \midrule
        MBPP  & Basic, SCoT, CoT   & Codex (Code-Davinci-002), GPT-3.5-Turbo     & SCoT \\
        \midrule
        MBCPP  & Basic, SCoT, CoT   & Codex (Code-Davinci-002), GPT-3.5-Turbo     & SCoT \\
    
\end{longtable}

\normalsize
\subsection{Free Response}
This task assess a model's ability in generating unconstrained textual response. The various datasets which we covered while reading up on different prompting methods for this task include Creative Writing \cite{yao2024tree} and Longform Generation of Biographies \cite{min2023factscore}. Table \ref{tab:freer} lists above-mentioned datasets and different prompting strategies that have been experimented on them along with the best technique. 
% prompting technique.

\footnotesize
\begin{longtable}{p{.6in}p{1.6in}p{2.6in}p{.5in}}
    \caption{Prompt Engineering Analysis for Free Response Task}
    \label{tab:freer} \\
    \toprule
    \textbf{Dataset}    & \textbf{Prompting Strategies} & \textbf{LLM(s)}  & \textbf{SoTA}  \\
    \midrule
    \endfirsthead
    
    \multicolumn{4}{c}%
    {{\bfseries \tablename\ \thetable{} -- continued from previous page}} \\
    \toprule
    \textbf{Dataset}    & \textbf{Prompting Strategies} & \textbf{LLM(s)}  & \textbf{SoTA}  \\
    \midrule
    \endhead

    \midrule \multicolumn{4}{r}{{Continued on next page}} \\
    \endfoot
    
    \bottomrule
    \endlastfoot

   Creative Writing   & Basic, CoT, Self-Consistency, ToT   & GPT-4      & ToT \\
        \midrule
        Longform Generation of Biographies  & Basic, CoT, CoVe    & LLaMA-65B, LLaMA-2-70B Chat     & CoVe \\
    
\end{longtable}

\normalsize

\subsection{Truthfulness}

This task assess a model's ability to communicate factually and not spread any kind of misinformation. This task does not represent a model's capability in understanding a given context, rather it is more focused on them not making false statements based on their understanding. The various datasets that we discovered while surveying different prompting strategies for this task are SycophancyEval, https://github.com/meg-tong/sycophancy-eval  \footnote{\url{https://github.com/meg-tong/sycophancy-eval}}, GSM-IC \cite{shi2023large} and Fever \cite{thorne2018fever}. Table \ref{tab:truth} shows above-mentioned datasets and different prompting techniques that have been experimented on them along with the best performing prompting technique.

\footnotesize
\begin{longtable}{p{.6in}p{1.6in}p{2.6in}p{.5in}}
    \caption{Prompt Engineering Analysis for Truthfulness Task}
    \label{tab:truth} \\
    \toprule
    \textbf{Dataset}    & \textbf{Prompting Strategies} & \textbf{LLM(s)}  & \textbf{SoTA}  \\
    \midrule
    \endfirsthead
    
    \multicolumn{4}{c}%
    {{ \tablename\ \thetable{} Continued from the Previous Page}} \\
    \midrule
    \textbf{Dataset}    & \textbf{Prompting Strategies} & \textbf{LLM(s)}  & \textbf{SoTA}  \\
    \midrule
    \endhead

     \multicolumn{4}{c}{{\tablename\ \thetable{} Continued on the Next Page}} \\
    \endfoot
    
    \bottomrule
    \endlastfoot

    Sycophancy-Eval  & S2A, CoT, Instructed Prompting    & LLaMA-2-70B-Chat      & S2A \\
        \midrule
        % GSM-IC  & S2A, CoT, Instructed Prompting    & LLaMA-2-70B-Chat      & 3493 \\
        % \midrule
        Longform Generation  & S2A, CoT, Instructed Prompting    & LLaMA-2-70B-Chat      & S2A \\
        \midrule
        Fever  & Basic, CoT, Act, ReAct, Self-Consistency, VE, CoK    & PaLM-540B, GPT-3.5 (Text-Davinci-002), GPT-3.5-Turbo, InstructGPT (Text-Davinci-003)     & ReAct \\
        \midrule
        % Fever  & Basic, CoT, Self-Consistency, VE, CoK    & GPT-3.5-Turbo     & 3493 \\
        % \midrule
        % GSM8K  & BadChain, DT-CoT, DT-Base    & GPT-4, GPT-3.5-Turbo, LLaMA-2-70B Chat, PaLM 2     & 3493 \\
        % \midrule
        % MATH  & BadChain, DT-CoT, DT-Base     & GPT-4, GPT-3.5-Turbo, LLaMA-2-70B Chat, PaLM 2     & 3493 \\
        % \midrule
        % ASDiv  & BadChain, DT-CoT, DT-Base     & GPT-4, GPT-3.5-Turbo, LLaMA-2-70B Chat, PaLM 2     & 3493 \\
        % \midrule
        % CSQA  & BadChain, DT-CoT, DT-Base     & GPT-4, GPT-3.5-Turbo, LLaMA-2-70B Chat, PaLM 2     & 3493 \\
        % \midrule
        % StrategyQA  & BadChain, DT-CoT, DT-Base     & GPT-4, GPT-3.5-Turbo, LLaMA-2-70B Chat, PaLM 2     & 3493 \\
        % \midrule
        % Last Letter Concatenation  & BadChain, DT-CoT, DT-Base    & GPT-4, GPT-3.5-Turbo, LLaMA-2-70B Chat, PaLM 2     & 3493 \\
        % \midrule
        GSM-IC  & CoT, Least-to-Most, Instructed Prompting, Self-Consistency, S2A    & Codex (Code-Davinci-002), GPT-3.5 ( Text-Davinci-003), LLaMA-2-70B-Chat     & Least-to-Most \\
    
\end{longtable}

\normalsize
\subsection{Table-Based Truthfulness}
 
This task is an extension of Truthfulness task and measures a model's ability to communicate factually and not spread any kind of misinformation in a tabular setting. The only dataset we came across while reading up on different prompting methods for this task is TabFact \cite{chen2019tabfact}. Table \ref{tab:tbtruth} contains above-mentioned datasets and different prompting strategies that have been experimented on them along with the best performing prompting strategy.

\footnotesize
\begin{longtable}{p{.6in}p{1.6in}p{2.6in}p{.5in}}
    \caption{Prompt Engineering Analysis for Table-Based Truthfulness Task}
    \label{tab:tbtruth} \\
    \toprule
    \textbf{Dataset}    & \textbf{Prompting Strategies} & \textbf{LLM(s)}  & \textbf{SoTA}  \\
    \midrule
    \endfirsthead
    
    \multicolumn{4}{c}%
    {{\bfseries \tablename\ \thetable{} -- continued from previous page}} \\
    \toprule
    \textbf{Dataset}    & \textbf{Prompting Strategies} & \textbf{LLM(s)}  & \textbf{SoTA}  \\
    \midrule
    \endhead

    \midrule \multicolumn{4}{r}{{Continued on next page}} \\
    \endfoot
    
    \bottomrule
    \endlastfoot

    TabFact  & Basic, CoT, Binder, Dater, Chain-of-Table    & PaLM 2-S, GPT-3.5-Turbo, LLaMA-2-17B-Chat     & Chain-of-Table \\

\end{longtable}

\normalsize
\subsection{Table-Based Question-Answering}

This task involves any kind of question-answering in a tabular setting. It can be considered as a super set of other kinds of table-based tasks like Table-Based Truthfulness or Table-Based Mathematical Problem Solving. But in this work, in order to avoid any confusion, we have captured all the datasets under this task which do not fall under more specific table-based tasks like Table-Based Truthfulness or Table-Based Mathematical Problem Solving. We came across only two datasets while reading up on different prompting strategies for this task which are FeTaQA \cite{nan2022fetaqa} and WikiTQ\cite{pasupat2015compositional}. Table \ref{tab:tbqa} shows above-mentioned datasets and different prompting methods that have been experimented on them along with the best performing prompting strategy.

\normalsize
\subsection{Table-Based Mathematical Problem Solving}

This task is an extension of Mathematical Problem Solving task and measures a model's ability to perform any kind of mathematical computation in a tabular setting. The different datasets which we covered while reading up on different prompting techniques for this task include TabMWP \cite{lu2022dynamic} and Penguins in a Table \cite{srivastava2022beyond}. Table \ref{tab:tbmps} lists above-mentioned datasets and different prompting methods that have been experimented on them along with the best performing prompting strategy.

\normalsize
\subsection{Recommender System}

This task measures a model's ability to process a given input and suggest a set of items which are most relevant from a list of possible items as output. We discovered only one dataset while surveying different prompting techniques for this task which is Movie Recommendation \cite{srivastava2022beyond}. Table \ref{tab:rec} lists above-mentioned datasets and different prompting methods that have been experimented on them along with the best performing prompting technique.

\footnotesize
\begin{longtable}{p{.6in}p{1.6in}p{2.6in}p{.5in}}
    \caption{Prompt Engineering Analysis for Table-Based Question-Answering Task}
    \label{tab:tbqa} \\
    \toprule
    \textbf{Dataset}    & \textbf{Prompting Strategies} & \textbf{LLM(s)}  & \textbf{SoTA}  \\
    \midrule
    \endfirsthead
    
    \multicolumn{4}{c}%
    {{ \tablename\ \thetable{} Continued from the Previous Page}} \\
    \toprule
    \textbf{Dataset}    & \textbf{Prompting Strategies} & \textbf{LLM(s)}  & \textbf{SoTA}  \\
    \midrule
    \endhead

    % \midrule 
    \multicolumn{4}{c}{{\tablename\ \thetable{} Continued on the Next Page}} \\
    \endfoot
    
    \bottomrule
    \endlastfoot

    WikiTQ  & Basic, CoT, Binder, Dater, Chain-of-Table    & PaLM 2-S, GPT-3.5-Turbo, LLaMA-2-17B-Chat, Codex (Code-Davinci-002)     & Chain-of-Table \\
        \midrule
        FeTaQA  & Basic, CoT, Dater, Chain-of-Table, Self-Consistency, VE, CoK    & PaLM 2-S, GPT-3.5-Turbo, LLaMA-2-17B-Chat, GPT-3.5-Turbo, Codex (Code-Davinci-002)     & Chain-of-Table \\
        % \midrule
        % FeTaQA  & Basic, CoT, Self-Consistency, VE, CoK    & GPT-3.5-Turbo     & 3493 \\

\end{longtable}

\footnotesize
\begin{longtable}{p{.6in}p{1.6in}p{2.6in}p{.5in}}
    \caption{Prompt Engineering Analysis for Table-Based Mathematical Problem Solving Task}
    \label{tab:tbmps} \\
    \toprule
    \textbf{Dataset}    & \textbf{Prompting Strategies} & \textbf{LLM(s)}  & \textbf{SoTA}  \\
    \midrule
    \endfirsthead
    
    \multicolumn{4}{c}%
    {{\bfseries \tablename\ \thetable{} -- continued from previous page}} \\
    \toprule
    \textbf{Dataset}    & \textbf{Prompting Strategies} & \textbf{LLM(s)}  & \textbf{SoTA}  \\
    \midrule
    \endhead

    \midrule \multicolumn{4}{r}{{Continued on next page}} \\
    \endfoot
    
    \bottomrule
    \endlastfoot

    TabMWP  & PoT, CoT, Self-Consistency, PAL    & Codex (Code-Davinci-002), GPT-3 (Text-Davinci-002), GPT-3.5-Turbo, CodeGen (Codegen-16B-Multi), CodeGen (Codegen-16B-Mono), CodeT5+, Xgen, PaLM, LaMDA    & PoT \\
        \midrule
        Penguins in a Table  & Basic, CoT, CoC, PAL, Random CoT, Complex CoT    & PaLM 2-S, GPT-3.5 (Text-Davinci-003), GPT-3.5-Turbo, GPT-4, Codex (Code-Davinci-002), UL2-20B, LaMDA-137B, PaLM-540B, Minerva-540B, GPT-3 (Text-Davinci-002), DiVeRSe      & PAL \\
        % \midrule
        % Penguins in a Table  & Basic, CoT, PAL    & Codex (Code-Davinci-002), UL2-20B, LaMDA-137B, PaLM-540B, Minerva-540B    & 3493 \\
        % \midrule
        % TabMWP  & Basic, CoT, PromptPG   & GPT-3    & 3493 \\
        % \midrule
        % Penguins in a Table  &  CoT, Random CoT, Complex CoT   & LaMDA-137B, PaLM-540B, Minerva-540B, GPT-3.5 (Text-Davinci-002), Codex (Code-Davinci-002), DiVeRSe    & 3493 \\

\end{longtable}

\footnotesize
\begin{longtable}{p{.6in}p{1.6in}p{2.6in}p{.5in}}
    \caption{Prompt Engineering Analysis for Recommender System Task}
    \label{tab:rec} \\
    \toprule
    \textbf{Dataset}    & \textbf{Prompting Strategies} & \textbf{LLM(s)}  & \textbf{SoTA}  \\
    \midrule
    \endfirsthead
    
    \multicolumn{4}{c}%
    {{\bfseries \tablename\ \thetable{} -- continued from previous page}} \\
    \toprule
    \textbf{Dataset}    & \textbf{Prompting Strategies} & \textbf{LLM(s)}  & \textbf{SoTA}  \\
    \midrule
    \endhead

    \midrule \multicolumn{4}{r}{{Continued on next page}} \\
    \endfoot
    
    \bottomrule
    \endlastfoot

    Movie Recommendation & Basic, CoT, CoC    & PaLM 2-S, GPT-3.5 (Text-Davinci-003), GPT-3.5-Turbo, GPT-4, Codex (Code-Davinci-002)    & Basic \\

\end{longtable}

\normalsize
\subsection{Emotion/Sentiment Understanding}

This task checks how good a model is at understanding human emotions or sentiments. The various datasets which we came across while reading up on different prompting methods for this task include Ruin Names \cite{srivastava2022beyond}, SemEval14 Laptop and Restaurant \cite{pontiki2016semeval} and Forex \cite{fatouros2023transforming}. Table \ref{tab:emotion} contains above-mentioned datasets and different prompting techniques that have been experimented on them along with the best performing prompting strategy.

\normalsize
\subsection{Machine Translation}

In this task, a model is tested on it's ability in terms of translation between two languages. The different datasets which we came across while reading up on different prompting techniques for this task include Salient Translation Error Detection \cite{srivastava2022beyond}, FLORES \cite{costa2022no}, WMT21 \cite{farhad2021findings},  Multi-Domain \cite{aharoni2020unsupervised} and PDC \cite{sun2020rethinking}. Table \ref{tab:mtrans} lists above-mentioned datasets and different prompting methods that have been experimented on them along with the best performing prompting strategy.

\normalsize
\subsection{Named Entity Recognition}

Named Entity Recognition task aims at identifying predefined classes or categories of objects in a given input text. The different datasets that we discovered while surveying different prompting techniques for this task are MTSamples \cite{uzuner20112010}, VAERS \cite{du2021extracting}, Research Papers \cite{tang2024large} and BC5CDR-chem \cite{li2016biocreative}. Table \ref{tab:ner} shows above-mentioned datasets and different prompting strategies that have been experimented on them along with the best performing prompting strategy.

\footnotesize
\begin{longtable}{p{.6in}p{1.6in}p{2.6in}p{.5in}}
    \caption{Prompt Engineering Analysis for Emotion/Sentiment Understanding Task}
    \label{tab:emotion} \\
    \toprule
    \textbf{Dataset}    & \textbf{Prompting Strategies} & \textbf{LLM(s)}  & \textbf{SoTA}  \\
    \midrule
    \endfirsthead
    
    \multicolumn{4}{c}%
    {{\bfseries \tablename\ \thetable{} -- continued from previous page}} \\
    \toprule
    \textbf{Dataset}    & \textbf{Prompting Strategies} & \textbf{LLM(s)}  & \textbf{SoTA}  \\
    \midrule
    \endhead

    \midrule \multicolumn{4}{r}{{Continued on next page}} \\
    \endfoot
    
    \bottomrule
    \endlastfoot

    Snarks & Basic, CoT, CoC    & PaLM 2-S, GPT-3.5 (Text-Davinci-003), GPT-3.5-Turbo, GPT-4    & CoC \\
        \midrule
        Ruin Names & Basic, CoT, CoC      & PaLM 2-S, GPT-3.5 (Text-Davinci-003), GPT-3.5-Turbo, GPT-4      & Basic \\
        \midrule
         SemEval14 Laptop and Restaurant   &  THOR, CoT     & Flan-T5-250M (Base), Flan-T5-780M (Large), Flan-T5-3B (XL), Flan-T5-11B (XXL), GPT3-350M, GPT3-1.3B, GPT3-6.7B, GPT3-175B, GPT-3.5-Turbo    & THOR \\
        \midrule
         Forex  & Basic, Basic + Variations     & GPT-3.5-Turbo     & Basic + Variations \\

\end{longtable}

\footnotesize
\begin{longtable}{p{.6in}p{1.6in}p{2.6in}p{.5in}}
    \caption{Prompt Engineering Analysis for Machine Translation Task}
    \label{tab:mtrans} \\
    \toprule
    \textbf{Dataset}    & \textbf{Prompting Strategies} & \textbf{LLM(s)}  & \textbf{SoTA}  \\
    \midrule
    \endfirsthead
    
    \multicolumn{4}{c}%
    {{ \tablename\ \thetable{} Continued from the Previous Page}} \\
    \midrule
    \textbf{Dataset}    & \textbf{Prompting Strategies} & \textbf{LLM(s)}  & \textbf{SoTA}  \\
    \midrule
    \endhead

    % \midrule 
    \multicolumn{4}{c}{{\tablename\ \thetable{} Continued on the Next Page}} \\
    \endfoot
    
    \bottomrule
    \endlastfoot

    Salient Translation Error Detection & Basic, CoT, CoC    & PaLM 2-S, GPT-3.5 (Text-Davinci-003), GPT-3.5-Turbo, GPT-4    & Basic \\
    \midrule
    FLORES & Basic, Basic + Variations     & GLM-130B      & Basic + Variations \\
        \midrule
        WMT21 & Basic, Basic + Variations     & GLM-130B      & Basic + Variations \\
        \midrule
         Multi-Domain  & Basic, Basic + Variations     & GLM-130B    & Basic + Variations \\
        \midrule
         PDC  & Basic, Basic + Variations     & GLM-130B     & Basic + Variations \\

\end{longtable}

% \normalsize
% \subsection{Named Entity Recognition}

% Named Entity Recognition task aims at identifying predefined classes or categories of objects in a given input text. The different datasets that we discovered while surveying different prompting techniques for this task are MTSamples \cite{uzuner20112010}, VAERS \cite{du2021extracting}, Research Papers \cite{tang2024large} and BC5CDR-chem \cite{li2016biocreative}. Table \ref{tab:ner} shows above-mentioned datasets and different prompting strategies that have been experimented on them along with the best performing prompting strategy.

\footnotesize
\begin{longtable}{p{.6in}p{1.6in}p{2.6in}p{.5in}}
    \caption{Prompt Engineering Analysis for Named Entity Recognition Task}
    \label{tab:ner} \\
    \toprule
    \textbf{Dataset}    & \textbf{Prompting Strategies} & \textbf{LLM(s)}  & \textbf{SoTA}  \\
    \midrule
    \endfirsthead
    
    \multicolumn{4}{c}%
    {{\bfseries \tablename\ \thetable{} -- continued from previous page}} \\
    \toprule
    \textbf{Dataset}    & \textbf{Prompting Strategies} & \textbf{LLM(s)}  & \textbf{SoTA}  \\
    \midrule
    \endhead

    \midrule \multicolumn{4}{r}{{Continued on next page}} \\
    \endfoot
    
    \bottomrule
    \endlastfoot

    MTSamples & Basic, Basic + Annotation Guideline-based Prompting, Basic + Annotation Guideline-Based Prompting + Error Analysis-Based Prompting    & GPT-3.5-Turbo, GPT-4     & Basic + Annotation Guideline-Based Prompting + Error Analysis-Based Prompting \\
        \midrule
        VAERS & Basic, Basic + Annotation Guideline-based Prompting, Basic + Annotation Guideline-Based Prompting + Error Analysis-Based Prompting     & GPT-3.5-Turbo, GPT-4      & Basic + Annotation Guideline-Based Prompting + Error Analysis-Based Prompting \\
        \midrule
        Research Papers & Basic, CoT     & GPT-3.5-Turbo, GPT-4     & Basic \\
        \midrule
        BC5CDR-chem & CoT, PS, Self-Consistency, MP     & Llama-2-13B-Chat, GPT-3.5-Turbo, GPT-4, PaLM-Bison-Chat     & MP \\

\end{longtable}

\normalsize
\subsection{Word Sense Disambiguation}

Word Sense Disambiguation task checks how good a model is at deciphering different meanings of a word in different contextual surroundings. We came across only one dataset while reading up on different prompting methods for this task which includes WiC \cite{pilehvar2018wic}. Table \ref{tab:wsd} shows above-mentioned datasets and different prompting techniques that have been experimented on them along with the best performing prompting method.

\normalsize
\subsection{Summarization}

This task tests a model's ability in breaking down a lengthy piece of input text into smaller chunks while ensuring retention of vital information in these smaller chunks. We covered only one dataset while reading up on different prompting methods for this task which is CCTC \cite{bao2024chain}. Table \ref{tab:summa} contains above-mentioned datasets and different prompting techniques that have been experimented on them along with the best performing prompting strategy.

\footnotesize
\begin{longtable}{p{.6in}p{1.6in}p{2.6in}p{.5in}}
    \caption{Prompt Engineering Analysis for Word Sense Disambiguation Task}
    \label{tab:wsd} \\
    \toprule
    \textbf{Dataset}    & \textbf{Prompting Strategies} & \textbf{LLM(s)}  & \textbf{SoTA}  \\
    \midrule
    \endfirsthead
    
    \multicolumn{4}{c}%
    {{\bfseries \tablename\ \thetable{} -- continued from previous page}} \\
    \toprule
    \textbf{Dataset}    & \textbf{Prompting Strategies} & \textbf{LLM(s)}  & \textbf{SoTA}  \\
    \midrule
    \endhead

    \midrule \multicolumn{4}{r}{{Continued on next page}} \\
    \endfoot
    
    \bottomrule
    \endlastfoot

    WiC & CoT, PS, Self-Consistency, MP     & Llama-2-13B-Chat, GPT-3.5-Turbo, GPT-4, PaLM-Bison-Chat     & MP \\

\end{longtable}

\footnotesize
\begin{longtable}{p{.6in}p{1.6in}p{2.6in}p{.5in}}
    \caption{Prompt Engineering Analysis for Summarization Task}
    \label{tab:summa} \\
    \toprule
    \textbf{Dataset}    & \textbf{Prompting Strategies} & \textbf{LLM(s)}  & \textbf{SoTA}  \\
    \midrule
    \endfirsthead
    
    \multicolumn{4}{c}%
    {{\bfseries \tablename\ \thetable{} -- continued from previous page}} \\
    \toprule
    \textbf{Dataset}    & \textbf{Prompting Strategies} & \textbf{LLM(s)}  & \textbf{SoTA}  \\
    \midrule
    \endhead

    \midrule \multicolumn{4}{r}{{Continued on next page}} \\
    \endfoot
    
    \bottomrule
    \endlastfoot

    WCEP & Basic, CoE     & ChatGLM2-6B      & CoE \\
        \midrule
        CCTC & Basic, CoE     & ChatGLM2-6B     & CoE \\

\end{longtable}

\normalsize
\subsection{Paraphrasing}

Paraphrasing task aims at rewriting a given piece of input text by using different words while keeping the true semantics of the original input text same. A key difference between Summarization task and Paraphrasing task is that the main goal of Summarization task is to shorten the length of output text with respect to that of input text whereas Paraphasing task just focuses on using different words during it's rewriting process. We discovered only one dataset while surveying different prompting methods for this task which includes QQP \footnote{\url{https://quoradata.quora.com/First-Quora-Dataset-Release-Question-Pairs}}. Table \ref{tab:parap} lists above-mentioned datasets and different prompting methods that have been experimented on them along with the best performing prompting technique.

\footnotesize
\begin{longtable}{p{.6in}p{1.6in}p{2.6in}p{.5in}}
    \caption{Prompt Engineering Analysis for Paraphrasing Task}
    \label{tab:parap} \\
    \toprule
    \textbf{Dataset}    & \textbf{Prompting Strategies} & \textbf{LLM(s)}  & \textbf{SoTA}  \\
    \midrule
    \endfirsthead
    
    \multicolumn{4}{c}%
    {{\bfseries \tablename\ \thetable{} -- continued from previous page}} \\
    \toprule
    \textbf{Dataset}    & \textbf{Prompting Strategies} & \textbf{LLM(s)}  & \textbf{SoTA}  \\
    \midrule
    \endhead

    \midrule \multicolumn{4}{r}{{Continued on next page}} \\
    \endfoot
    
    \bottomrule
    \endlastfoot

    QQP & CoT, PS, Self-Consistency, MP     & Llama-2-13B-Chat, GPT-3.5-Turbo, GPT-4, PaLM-Bison-Chat     & MP \\

\end{longtable}

\normalsize
\subsection{Stance Detection}

This task evaluates a model's ability in determining  from text whether the author of the text is in favor or against a topic or target or an object of evaluation. The different datasets which we came across while reading up on different prompting techniques for this task are SemEval-2016 \cite{mohammad2016semeval}, VAST \cite{allaway2020zero} and P-Stance \cite{li2021p}. Table \ref{tab:stanced} shows above-mentioned datasets and different prompting methods that have been experimented on them along with the best performing prompting technique.

\footnotesize
\begin{longtable}{p{.6in}p{1.6in}p{2.6in}p{.5in}}
    \caption{Prompt Engineering Analysis for Stance Detection Task}
    \label{tab:stanced} \\
    \toprule
    \textbf{Dataset}    & \textbf{Prompting Strategies} & \textbf{LLM(s)}  & \textbf{SoTA}  \\
    \midrule
    \endfirsthead
    
    \multicolumn{4}{c}%
    {{\bfseries \tablename\ \thetable{} -- continued from previous page}} \\
    \toprule
    \textbf{Dataset}    & \textbf{Prompting Strategies} & \textbf{LLM(s)}  & \textbf{SoTA}  \\
    \midrule
    \endhead

    \midrule \multicolumn{4}{r}{{Continued on next page}} \\
    \endfoot
    
    \bottomrule
    \endlastfoot

    SemEval-2016 &  CoT     & GPT-3.5-Turbo     & CoT \\
        \midrule
        VAST &  CoT     & GPT-3.5-Turbo      & CoT \\
        \midrule
        P-Stance &  CoT    & GPT-3.5-Turbo      & CoT \\

\end{longtable}

\normalsize
\subsection{Natural Language Inference}

The main objective of this task is to determine whether a \textit{hypothesis} is true (entailment), false (contradiction), or undetermined (neutral) given a \textit{premise}. The different datasets which we covered while reading up on different prompting methods for this task are QNLI \cite{rajpurkar2016squad} and MedNLI \cite{romanov2018lessons}. Table \ref{tab:nli} contains above-mentioned datasets and different prompting strategies that have been experimented on them along with the best performing prompting method.

\footnotesize
\begin{longtable}{p{.6in}p{1.6in}p{2.6in}p{.5in}}
    \caption{Prompt Engineering Analysis for Natural Language Inference Task}
    \label{tab:nli} \\
    \toprule
    \textbf{Dataset}    & \textbf{Prompting Strategies} & \textbf{LLM(s)}  & \textbf{SoTA}  \\
    \midrule
    \endfirsthead
    
    \multicolumn{4}{c}%
    {{\bfseries \tablename\ \thetable{} -- continued from previous page}} \\
    \toprule
    \textbf{Dataset}    & \textbf{Prompting Strategies} & \textbf{LLM(s)}  & \textbf{SoTA}  \\
    \midrule
    \endhead

    \midrule \multicolumn{4}{r}{{Continued on next page}} \\
    \endfoot
    
    \bottomrule
    \endlastfoot

    QNLI & CoT, PS, Self-Consistency, MP     & Llama-2-13B-Chat, GPT-3.5-Turbo, GPT-4, PaLM-Bison-Chat     & MP \\
        \midrule
        MedNLI & CoT, PS, Self-Consistency, MP     & Llama-2-13B-Chat, GPT-3.5-Turbo, GPT-4, PaLM-Bison-Chat     & MP \\

\end{longtable}

\normalsize
\subsection{Relation Extraction}

Relation Extraction evaluates a model's ability in identifying semantic relationships between predefined classes or categories of objects or named entities. We came across only one dataset while reading up on different prompting techniques for this task which includes DDI \cite{segura2013semeval}. Table \ref{tab:re} shows above-mentioned datasets and different prompting methods that have been experimented on them along with the best performing prompting strategy.
\footnotesize
\begin{longtable}{p{.6in}p{1.6in}p{2.6in}p{.5in}}
    \caption{Prompt Engineering Analysis for Relation Extraction Task}
    \label{tab:re} \\
    \toprule
    \textbf{Dataset}    & \textbf{Prompting Strategies} & \textbf{LLM(s)}  & \textbf{SoTA}  \\
    \midrule
    \endfirsthead
    
    \multicolumn{4}{c}%
    {{\bfseries \tablename\ \thetable{} -- continued from previous page}} \\
    \toprule
    \textbf{Dataset}    & \textbf{Prompting Strategies} & \textbf{LLM(s)}  & \textbf{SoTA}  \\
    \midrule
    \endhead

    \midrule \multicolumn{4}{r}{{Continued on next page}} \\
    \endfoot
    
    \bottomrule
    \endlastfoot

    DDI & CoT, PS, Self-Consistency, MP     & Llama-2-13B-Chat, GPT-3.5-Turbo, GPT-4, PaLM-Bison-Chat     & MP \\

\end{longtable}

% \begin{table*}[h]
%     \centering
%     \scalebox{0.85}{
%     \begin{tabular}{p{.6in}p{1.6in}p{3.0in}p{.5in}}
%     \toprule
%         \textbf{Dataset}    & \textbf{Prompting Strategies} & \textbf{LLM(s)}  & \textbf{SoTA}  \\
%         \midrule
%         WikiTQ & Basic, CoT, Text-to-SQL, Binder, Dater, Chain-of-Table & PaLM 2-S, GPT-3.5-Turbo, LLaMA-2-17B-Chat     & 3493 \\
%         \midrule
%         FeTaQA & Basic, CoT, Text-to-SQL, Binder, Dater, Chain-of-Table, Self-Consistency, VE, CoK  & PaLM 2-S, GPT-3.5-Turbo, LLaMA-2-17B-Chat    & 3493 \\
%     \end{tabular}}
%     \caption{Prompt Engineering Analysis for Table-Based Question-Answering Task}
%     \label{tab:tabbasedqa}
% \end{table*}

% \begin{table*}[h]
%     \centering
%     \scalebox{0.85}{
%     \begin{tabular}{p{.6in}p{1.6in}p{3.0in}p{.5in}}
%     \toprule
%         \textbf{Dataset}    & \textbf{Prompting Strategies} & \textbf{LLM(s)}  & \textbf{SoTA}  \\
%         \midrule
%         TabFact & Basic, CoT, Text-to-SQL, Binder, Dater, Chain-of-Table & PaLM 2-S, GPT-3.5-Turbo, LLaMA-2-17B-Chat     & 3493 \\
%     \end{tabular}}
%     \caption{Prompt Engineering Analysis for Table-Based Truthfulness Task}
%     \label{tab:tabbasedtruth}
% \end{table*}

\normalsize
\subsection{Language-Based Task Completion}

The main objective of this task to check how good is a model in following a sequence of language-based navigational commands to make decisions about it's actions required to complete a task.The different datasets that we discovered while surveying different prompting strategies for this task are ALFWorld \cite{shridhar2020alfworld}, WebShop \cite{yao2022webshop}, SayCan \cite{ahn2022can} and Scan \cite{lake2018generalization}. Table \ref{tab:langbased} lists above-mentioned datasets and different prompting methods that have been experimented on them along with the best performing prompting method.

\footnotesize
\begin{longtable}{p{.6in}p{1.6in}p{2.6in}p{.5in}}
    \caption{Prompt Engineering Analysis for Language-Based Task Completion Task}
    \label{tab:langbased} \\
    \toprule
    \textbf{Dataset}    & \textbf{Prompting Strategies} & \textbf{LLM(s)}  & \textbf{SoTA}  \\
    \midrule
    \endfirsthead
    
    \multicolumn{4}{c}%
    {{ \tablename\ \thetable{} Continued from the Previous Page}} \\
    % \toprule
    \midrule
    \textbf{Dataset}    & \textbf{Prompting Strategies} & \textbf{LLM(s)}  & \textbf{SoTA}  \\
    \midrule
    \endhead

    \multicolumn{4}{c}{{\tablename\ \thetable{} Continued on the Next Page}} \\
    \endfoot
    
    \bottomrule
    \endlastfoot

    ALFWorld & Act, ReAct     & PaLM-540B, GPT-3 (Text-Davinci-002)     & ReAct \\
        \midrule
        Scan    & Basic, CoT, Least-to-Most    & GPT-3 (Text-Davinci-002), Codex (Code-Davinci-001), Codex (Code-Davinci-001)  & Least-to-Most\\
        \midrule
        WebShop    & Act, ReAct     & PaLM-540B, GPT-3 (Text-Davinci-002)   & ReAct  \\
        \midrule
        
        SayCan    & Basic, CoT    & GPT-3 (Text-Davinci-002), LaMDA-137B, PaLM-540B, UL2-20B, Codex (Code-Davinci-002)  & CoT\\

\end{longtable}

\normalsize
\subsection{Multilabel Text Classification}

This task measures a model's ability to assign each input to a set of predefined target labels. This task can encapsulate a lot of above-mentioned tasks like Stance Detection, Named Entity Recognition etc but again in order to keep these task definitions as disjoint as possible for a better survey of prompting methods, we have included only those datasets under this task which could not be suitably categorized under any of the above-discussed tasks. The different datasets which we covered while reading up on different prompting strategies for this task include EUR-LEX \cite{chalkidis2021multieurlex}, UNFAIR-ToS \cite{lippi2019claudette} and LEDGAR \cite{tuggener2020ledgar}. Table \ref{tab:mlc} contains above-mentioned datasets and different prompting strategies that have been experimented on them along with the best performing prompting method.

\footnotesize
\begin{longtable}{p{.6in}p{1.6in}p{2.6in}p{.5in}}
    \caption{Prompt Engineering Analysis for Multilabel Text Classification Task}
    \label{tab:mlc} \\
    \toprule
    \textbf{Dataset}    & \textbf{Prompting Strategies} & \textbf{LLM(s)}  & \textbf{SoTA}  \\
    \midrule
    \endfirsthead
    
    \multicolumn{4}{c}%
    {{\bfseries \tablename\ \thetable{} -- continued from previous page}} \\
    \toprule
    \textbf{Dataset}    & \textbf{Prompting Strategies} & \textbf{LLM(s)}  & \textbf{SoTA}  \\
    \midrule
    \endhead

    \midrule \multicolumn{4}{r}{{Continued on next page}} \\
    \endfoot
    
    \bottomrule
    \endlastfoot

    EUR-LEX & CoT, PS, Self-Consistency, MP     & Llama-2-13B-Chat, GPT-3.5-Turbo, GPT-4, PaLM-Bison-Chat     & MP \\
        \midrule
        UNFAIR-ToS    & CoT, PS, Self-Consistency, MP     & Llama-2-13B-Chat, GPT-3.5-Turbo, GPT-4, PaLM-Bison-Chat   & MP  \\
        \midrule
        LEDGAR     & CoT, PS, Self-Consistency, MP    & Llama-2-13B-Chat, GPT-3.5-Turbo, GPT-4, PaLM-Bison-Chat  & MP\\

\end{longtable}

\normalsize
\section{Conclusion}

Prompt engineering has become indispensable in the present realm of LLMs. It plays a crucial role in realising the full potential of LLMs through various measures. In this work, we do an in-depth survey of 44 research papers talking about 39 prompting strategies across 29 different NLP tasks. We pictorially present this through a taxonomy diagram. We try to standardize the categorization of different datasets into 29 NLP tasks and discuss the overall effect of recent prompting techniques across them while also listing down potential SoTA prompting method for each dataset.

\bibliography{iclr2024_conference}
\bibliographystyle{iclr2024_conference}

% \appendix
% \section{Appendix}

\end{document}